\documentclass[sn-mathphys,Numbered,iicol]{sn-jnl}

\usepackage{graphicx}
\usepackage{multirow}
\usepackage{array}

\usepackage{amsmath,amssymb,amsfonts}
\usepackage{amsthm}
\usepackage{mathrsfs}
\usepackage[title]{appendix}
\usepackage{xcolor}%
\usepackage{textcomp}%
\usepackage{manyfoot}%

\usepackage{algorithm}%
\usepackage{algorithmicx}%
\usepackage{algpseudocode}%
\usepackage{listings}%
\usepackage{tabularx}%

\usepackage{ragged2e}

\theoremstyle{thmstyleone}%
\theoremstyle{thmstyletwo}%

\theoremstyle{thmstylethree}%

\providecommand{\new}[1]{\textcolor{black}{#1}}

\begin{document}

\title[Article Title]{Data Augmentation in Human-Centric Vision}


\author*[1]{\fnm{Wentao} \sur{Jiang}}\email{jiangwentao@buaa.edu.cn}

\author[1]{\fnm{Yige} \sur{Zhang}}\email{19241002@buaa.edu.cn}

\author[1]{\fnm{Shaozhong} \sur{Zheng}}\email{19241094@buaa.edu.cn}

\author[1]{\fnm{Si} \sur{Liu}} \email{liusi@buaa.edu.cn}

\author[2]{\fnm{Shuicheng} \sur{Yan}}\email{shuicheng.yan@kunlun-inc.com}

\affil[1]{\orgname{Beihang University}, \orgaddress{\street{Xueyuan Road No.37}, \city{Beijing}, \postcode{100191}, \country{China}}}


\affil[2]{ \orgname{Skywork AI}, \orgaddress{\street{7 Straits View}, \city{Singapore}, \postcode{018936}, \country{Singapore}}}



\abstract{
    This survey presents a comprehensive analysis of data augmentation techniques in human-centric vision tasks, a first of its kind in the field. It delves into a wide range of research areas including person ReID, human parsing, human pose estimation, and pedestrian detection, addressing the significant challenges posed by overfitting and limited training data in these domains. Our work categorizes data augmentation methods into two main types: data generation and data perturbation. Data generation covers techniques like graphic engine-based generation, generative model-based generation, and data recombination, while data perturbation is divided into image-level and human-level perturbations. Each method is tailored to the unique requirements of human-centric tasks, with some applicable across multiple areas. Our contributions include an extensive literature review, providing deep insights into the influence of these augmentation techniques in human-centric vision and highlighting the nuances of each method. We also discuss open issues and future directions, such as the integration of advanced generative models like Latent Diffusion Models, for creating more realistic and diverse training data. This survey not only encapsulates the current state of data augmentation in human-centric vision but also charts a course for future research, aiming to develop more robust, accurate, and efficient human-centric vision systems.
}




\maketitle

\section{Introduction}\label{sec1}

Human-centric perception remains a core focus within the computer vision and machine learning communities, encompassing a wide array of research tasks and applications such as person ReID~\cite{ye2021deep,zheng2016person,zheng2015scalable,wang2020comprehensive,zheng2017person,wu2019deep,zhang2017alignedreid}, human parsing~\cite{li2020self,ruan2019devil,gong2018instance,li2017multiple,gong2019graphonomy,wang2020hierarchical,gong2017look}, human pose estimation~\cite{fang2017rmpe,chen20173d,martinez2017simple,andriluka2018posetrack,papandreou2017towards,belagiannis2017recurrent,zheng2023deep,zheng20213d}, and pedestrian detection~\cite{liu2019high,gawande2020pedestrian,mao2017can,zhang2017citypersons,lan2018pedestrian,iftikhar2022deep,roszyk2022adopting}. Despite significant advancements, these algorithms are inherently data-intensive and frequently encounter overfitting issues~\cite{ying2019overview,bejani2021systematic,bartlett2020benign,chen2020robust,zhang2021lbcf,hosseini2020tried,zhang2022towards}, where models excel with training data but falter on unseen test data. This challenge intensifies when access to large datasets is limited, often due to privacy concerns or the need for labor-intensive and costly human annotation tasks~\cite{ashktorab2021ai}.

Addressing this issue, data augmentation emerges as a practical solution, particularly in the context of high collection and annotation costs. While previous surveys~\cite{shorten2019survey,kumar2023advanced} have explored data augmentation across various computer vision tasks, they often overlook the unique aspects of human-centric vision tasks. This survey seeks to fill this gap by extensively summarizing data augmentation works specific to human-centric vision tasks, such as person ReID, human parsing, human pose estimation, and pedestrian detection. These tasks, while distinct, share commonalities in augmentation methods that leverage human-specific features, with some techniques being applicable across multiple human-centric tasks.

We aim to comprehensively survey data augmentation methods in human-centric vision, including several works previously discussed in surveys~\cite{shorten2019survey,kumar2023advanced}, while primarily focusing on methods not extensively covered before, especially those concerning human body augmentation. Data augmentation techniques in this domain can be broadly categorized into two types: data generation and data perturbation. Data generation involves creating or expanding datasets by adding new examples through various techniques, such as collecting additional samples, applying transformations, or introducing variations to bolster model training and generalization. This category encompasses graphic engine-based generation, generative model-based generation, and data recombination.

Complementing data generation, data perturbation is frequently employed, subdivided into image-level and human-level perturbations. Image-level perturbation involves applying transformations to the entire image, such as rotations, flips, zooms, or adjustments in brightness and contrast, to artificially enhance the training dataset's diversity. On the other hand, human-level perturbation introduces alterations at the level of individual samples, aiming to increase a model's adaptability to varied instances by simulating the diversity encountered in real scenes.

Furthermore, data augmentation methods can be differentiated based on their application in specific human-centric tasks, including person ReID, human parsing, human pose estimation, and pedestrian detection. Each of these tasks employs distinct augmentation methods that perform both data perturbation and data generation, which will be elaborated upon in this survey.

Our contribution can be summarized as:
\begin{itemize}
    \item We are the first to conduct a comprehensive survey of data augmentation methods tailored for human-centric vision tasks, highlighting the unique characteristics of these methods in relation to human-centric tasks.
    \item We provide a comprehensive literature review on data augmentation methods for human-centric vision tasks, summarizing and categorizing methods from various perspectives. This provides an in-depth understanding of crucial factors influencing augmentation techniques in human-centric vision.
    \item We present a thorough discussion about open issues and potential future directions based on our investigation. Our comprehensive studies uncover the pros/cons of current methods and bring new observations and insights to the community.
\end{itemize}

\begin{table*}[]
\renewcommand{\arraystretch}{1.5}
\scalebox{0.76}{
\begin{tabular}{|c|cc|l|}
\hline
\textbf{Categories} & \multicolumn{1}{c|}{\textbf{Sub-categories}} & \textbf{\begin{tabular}[c]{@{}c@{}}Subsub-\\ categories\end{tabular}} & \multicolumn{1}{c|}{\textbf{Methods}} \\ \hline
\multirow{12}{*}{\textbf{\begin{tabular}[c]{@{}c@{}}Data\\ perturbation\end{tabular}}} & \multicolumn{1}{c|}{\multirow{6}{*}{\textbf{\begin{tabular}[c]{@{}c@{}}Image-level\\perturbation\end{tabular}}}} & \multirow{3}{*}{\textbf{\begin{tabular}[c]{@{}c@{}}Global\\ perturbation\end{tabular}}} & \textbf{Scaling and rotation:} (X Peng 2018~\cite{peng2018jointly}) \\ \cline{4-4} 
 & \multicolumn{1}{c|}{} &  & \begin{tabular}[c]{@{}l@{}}\textbf{Style transfer:} (CamStyle 2018~\cite{zhong2018camstyle})(C Michaelis 2019~\cite{michaelis2019benchmarking})\\ (Z Zhong 2018~\cite{zhong2018camera})(Z lin 2021~\cite{lin2021color})\end{tabular} \\ \cline{4-4} 
 & \multicolumn{1}{c|}{} &  & \textbf{Noise injection:} (Wang 2021~\cite{wang2021human}) \\ \cline{3-4} 
 & \multicolumn{1}{c|}{} & \multirow{3}{*}{\textbf{\begin{tabular}[c]{@{}c@{}}Region-level\\ perturbation\end{tabular}}} & \begin{tabular}[c]{@{}l@{}}\textbf{Information dropping:} (Z Zhong 2020~\cite{zhong2020random})(J Huang 2020~\cite{huang2020aid})\\ (W Sun 2020~\cite{sun2020triplet})(Pedhunter 2020~\cite{chi2020pedhunter})\end{tabular} \\ \cline{4-4} 
 & \multicolumn{1}{c|}{} &  & \textbf{Grayscale patch:} (Y Gong 2021~\cite{Gong2021APR}) \\ \cline{4-4} 
 & \multicolumn{1}{c|}{} &  & \textbf{Patch stylized:} (S Cygert 2020~\cite{Cygert2020TowardRP}) \\ \cline{2-4} 
 & \multicolumn{1}{c|}{\multirow{6}{*}{\textbf{\begin{tabular}[c]{@{}c@{}}Human-level\\perturbation\end{tabular}}}} & \multirow{3}{*}{\textbf{\begin{tabular}[c]{@{}c@{}}Human-level\\ occlusion\\ generation\end{tabular}}} & \textbf{Keypoint masking:} (L Ke 2018~\cite{ke2018multi}) \\ \cline{4-4} 
 & \multicolumn{1}{c|}{} &  & \textbf{Copy-paste for complex scenario synthesis:} (Y Bin 2020~\cite{bin2020adversarial}) \\ \cline{4-4} 
 & \multicolumn{1}{c|}{} &  & \textbf{Nearby-person occlusion:} (Y Chen 2021~\cite{chen2021nearby}) \\ \cline{3-4} 
 & \multicolumn{1}{c|}{} & \multirow{3}{*}{\textbf{\begin{tabular}[c]{@{}c@{}}Human body\\ perturbation\end{tabular}}} & \textbf{Altering pedestrian shapes:} (Zh Chen 2019~\cite{chen2021shape}) \\ \cline{4-4} 
 & \multicolumn{1}{c|}{} &  & \textbf{2D human pose transformation:} (PoseTrans 2022~\cite{jiang2022posetrans}) \\ \cline{4-4} 
 & \multicolumn{1}{c|}{} &  & \begin{tabular}[c]{@{}l@{}}\textbf{3D human pose transformation:} (Li 2021~\cite{li2020cascaded})(Z Xin 2022~\cite{xin20223d})\\ (L Huang 2022~\cite{huang2022dh})(PoseGU 2023~\cite{guan2023posegu})\end{tabular} \\ \hline
\multirow{6}{*}{\textbf{\begin{tabular}[c]{@{}c@{}}Data\\ generation\end{tabular}}} & \multicolumn{2}{c|}{\textbf{\begin{tabular}[c]{@{}c@{}}Graphic\\ Engine-based\end{tabular}}} & \begin{tabular}[c]{@{}l@{}}(MixedPeds 2017~\cite{Cheung2017MixedPedsPD})(W Chen 2017~\cite{chen2016synthesizing})(Varol 2018~\cite{varol2017learning})\\ (Bo lu 2022~\cite{Lu2022PedestrianDF})(SynPoses 2022~\cite{nie2022synposes})(J Nilsson 2014~\cite{nilsson2014pedestrian})\\(D Mehta 2018~\cite{mehta2017vnect})\end{tabular} \\ \cline{2-4} 
 & \multicolumn{2}{c|}{\textbf{\begin{tabular}[c]{@{}c@{}}Generative\\ Model-based\end{tabular}}} & \begin{tabular}[c]{@{}l@{}}(A Siarohin 2018~\cite{siarohin2018deformable})(X Zhang 2021~\cite{zhang2020deep})(PAC-GAN 2020~\cite{zhang2020pac})\\(S Liu 2020~\cite{liu2020novel})(FD-GAN 2018~\cite{Ge2018FDGANPF})(L Zhang 2021~\cite{Zhang2021PoseVA})\\(V Uc-Cetina 2023~\cite{uc2023review})(Z Yang 2023~\cite{yang2023improved})(J Liu 2018~\cite{Liu2018PoseTP})\\(R Zhi 2021~\cite{Zhi2021PoseGuidedPI})(D Wu 2018~\cite{wu2018random})(Q Wu 2021~\cite{wu2021deep})\end{tabular} \\ \cline{2-4} 
 & \multicolumn{1}{c|}{\multirow{4}{*}{\textbf{\begin{tabular}[c]{@{}c@{}}Data\\ recombination\end{tabular}}}} & \multirow{2}{*}{\textbf{\begin{tabular}[c]{@{}c@{}}Image-level\\ recombination\end{tabular}}} & \begin{tabular}[c]{@{}l@{}}\textbf{Background replacing:} (N McLaughlin 2015~\cite{McLaughlin2015DataaugmentationFR})(Dai 2022~\cite{dai2023overcoming})\\ (T Kikuchi 2017~\cite{Kikuchi2018TransferringPA})(L Chen 2017~\cite{Chen2017DataGF})(M Tian 2018~\cite{tian2018eliminating})\end{tabular} \\ \cline{4-4} 
 & \multicolumn{1}{c|}{} &  & \begin{tabular}[c]{@{}l@{}}\textbf{Copy-paste:} (D Dwibedi 2017~\cite{dwibedi2017cut})(CL Li 2021~\cite{li2021cutpaste})\\(Instaboost 2019~\cite{fang2019instaboost})(J Deng 2022~\cite{Deng2022ImprovingCO})\\(G Ghiasi 2020~\cite{ghiasi2021simple})(T Remez 2018~\cite{Remez2018LearningTS})\end{tabular} \\ \cline{3-4} 
 & \multicolumn{1}{c|}{} & \multirow{2}{*}{\textbf{\begin{tabular}[c]{@{}c@{}}Human-level\\ recombination\end{tabular}}} & \textbf{2D image:} (F Chen 2020~\cite{chen2020self})(K Han 2023~\cite{han2023clothing})(X Jia 2022~\cite{jia2022complementary}) \\ \cline{4-4} 
 & \multicolumn{1}{c|}{} &  & \textbf{3D pose:} (PoseAug 2022~\cite{gong2021poseaug}) \\ \hline
\end{tabular}}
\vspace{2mm}
\caption{Categorized by data augmentation methods.}
\label{tab:category1}
\end{table*}

\textbf{Outline.}
The remainder of this paper is organized as follows. Section~\ref{sec2} introduces the existing review papers classified by the type of augmentation method.
Sections~\ref{sec3} describe augmentation methods categorized by the applied human-centric tasks.
Finally, Section~\ref{sec4} concludes the paper and discusses several promising future research directions.

\section{Categorized by Data Augmentation Method}
\label{sec2}

The proposed taxonomy classifies data augmentation in human-centric vision into two main branches: data perturbation and data augmentations, as presented in Table~\ref{tab:category1}.
The former indicates methods that perturb the original example for data augmentation, while the latter refers to methods that generate training new examples for data augmentation.
The specifics of each data augmentation method are thoroughly discussed in subsequent sections.

\subsection{Data Perturbation}

Data perturbation aims to augment data using the existing original example. It can be classified as image-level data perturbation that applies transformation, erasing, and mixing in the whole image and human-level data perturbation that only alters the human instances.

\subsubsection{Image-level Perturbation}

Image-level data augmentation method involves applying various transformations to the image to artificially increase the diversity of the training dataset. These transformations may include rotations, flips, zooms, or brightness changes in image-level or region-level. By augmenting the dataset with these modified images, the model becomes more robust and better generalizes to variations in the input data, enhancing its performance in tasks related to human-centric vision.

\begin{figure*}[t]
  \centering
  \includegraphics[width=0.9\linewidth]{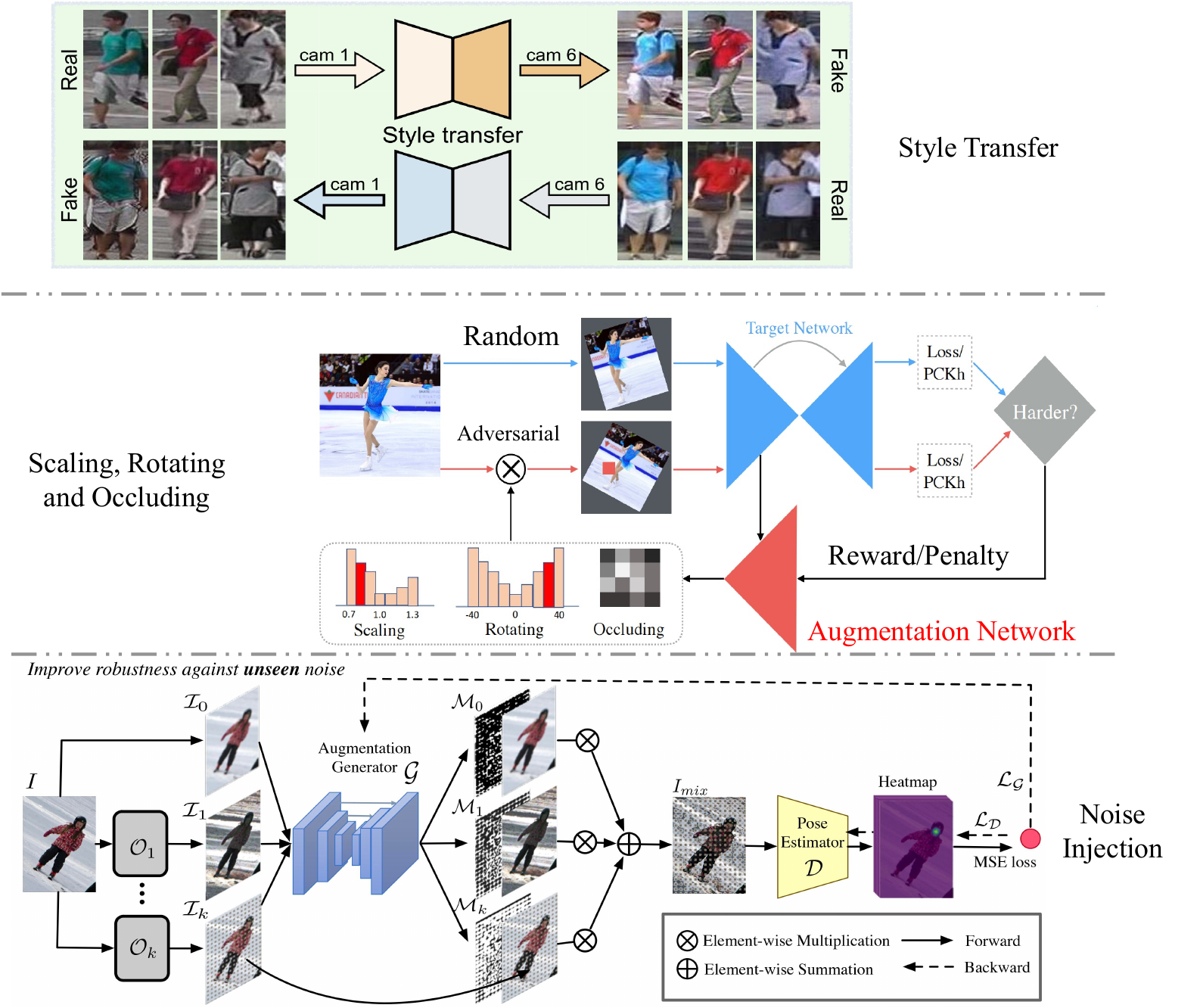}
  \caption{
    Examples of global perturbation. The figure contains representative works of Style Transfer~\cite{zhong2018camstyle}, Scaling Rotating and Occluding~\cite{peng2018jointly} and Noise Injection~\cite{wang2021human}.
  }
  \label{global_per}
\end{figure*}

\textbf{Global Perturbation.}
Global perturbation methods modify the overall characteristics of an image, introducing variations such as noise addition~\cite{wang2021human} or style transfer~\cite{zhong2018camstyle}, as shown in Figure~\ref{global_per}. These techniques aim to change the global appearance of the image, providing the model with exposure to different visual patterns and scenarios during training.
    \begin{itemize}
        \item Style Transfer~\cite{zhong2018camstyle,zhong2018camera,michaelis2019benchmarking,lin2021color}: In the realm of global perturbation methods, one notable technique involves the stylization of training images. This method enhances robustness against various types of corruptions, severities, and across different datasets. It is achieved by blending the content of an image with the style elements of another, leading to a significant improvement in the model's ability to generalize across diverse visual scenarios. This technique underscores the value of style variability in training data for enhancing model robustness.
        \item Scaling, Rotating, and Occluding~\cite{peng2018jointly}: Another approach in global perturbations is the use of an augmentation network designed to create adversarial distributions. From these distributions, specific augmentation operations such as scaling, rotating, and occluding are sampled to generate novel data points. This method represents a strategic shift towards adversarial robustness, where the model is trained on data points that are systematically varied to challenge and thereby improve the model's resilience and adaptability to real-world variations in visual data.
        \item Noise Injection~\cite{wang2021human}: AdvMix~\cite{huang2021advmix} exemplifies a noise injection approach to improve the robustness of human pose estimation models against data corruptions. This method, adaptable across various human pose estimation frameworks, employs a combination of adversarial augmentation to introduce challenging corrupted images and knowledge distillation to preserve clean pose information. This strategy effectively trains models to withstand a range of data inconsistencies, enhancing their real-world applicability.
    \end{itemize}

\begin{figure*}[!t]
  \centering
  \includegraphics[width=0.9\linewidth]{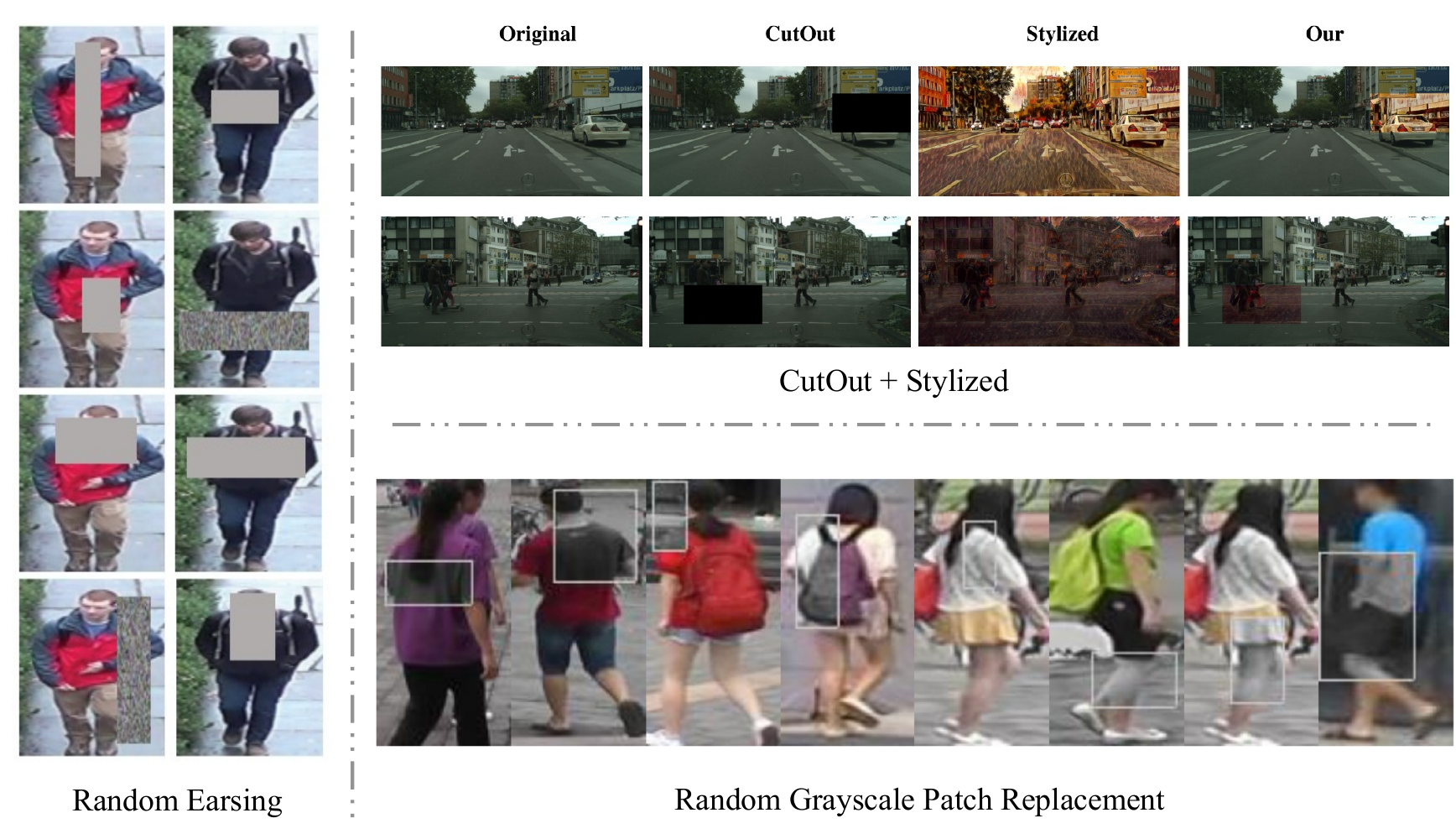}
  \caption{
  Examples of region-level perturbation. The figure contains representative works of Random Earsing~\cite{zhong2020random}, CutOut and Stylized~\cite{Cygert2020TowardRP} and Random Grayscale Patch Replacement~\cite{Gong2021APR}. }
  \label{region_per}
\end{figure*}

\new{Global perturbation methods in data augmentation present notable advantages and drawbacks. On the positive side, their uniform application across all data instances simplifies operations, contributing to ease of implementation and potentially enhancing the model's overall robustness. This simplicity is particularly advantageous in ensuring consistent treatment of diverse data, simplifying the training pipeline, and making the model more adaptable to varying scenarios.}

\new{However, these advantages come with certain caveats. Excessive modifications through global perturbation may distort original features, impacting the model's ability to discern meaningful patterns. The risk of overemphasis on artificial variations poses a challenge, potentially diminishing the model's efficacy in recognizing significant features representative of the natural data distribution. Furthermore, the selection of appropriate perturbation techniques requires careful consideration. Inappropriate transformations may introduce unrealistic scenarios, leading to suboptimal model performance by deviating from real-world data characteristics.}

\textbf{Region-level Perturbation.}
Augmentation methods, as shown in Figure~\ref{region_per},  involving region-level information alteration focus on introducing variability by randomly manipulating specific areas of an image. Techniques like random erasing~\cite{zhong2020random} or adding random grayscale patches~\cite{Gong2021APR} selectively modify portions of the image, contributing to a more diverse training set and improving the model's ability to handle localized variations.
\begin{itemize}
    \item Information Dropping: This is a category of data augmentation techniques that includes methods like random erasing~\cite{zhong2020random}, Cutout, and Grid Mask~\cite{huang2020aid}. These methods enhance model training by deliberately removing or obscuring parts of an image. random erasing randomly selects and erases a rectangular region in an image, Cutout removes specific patches, and Grid Mask applies a grid-like pattern to mask parts of the image. All these techniques introduce varying levels of occlusion or information loss, effectively simulating real-world scenarios where parts of a scene may be obscured or missing.
    \item Cutout with Stylized Patch Integration~\cite{Cygert2020TowardRP}: The cutout method, inspired by the concept of patch Gaussian augmentation, adds a patch of a stylized image to the original image at a randomly sampled location. This technique allows for variation in the size of the patch, introducing a unique form of augmentation. By selectively applying stylized patches, Cutout improves the model's accuracy and its robustness. This method simulates scenarios where part of the human subject or the background may undergo unusual visual changes, challenging the model to maintain performance under diverse conditions.
    \item Random Grayscale Patch Replacement~\cite{Gong2021APR}: This method involves selectively altering regions within an image by replacing them with corresponding grayscale patches. The process, encompassing Local Grayscale Patch Replacement (LGPR) and Global Grayscale Patch Replacement (GGPR), introduces varying levels of grayscale into the training images. By doing so, Random Grayscale Patch Replacement provides a unique form of region-level perturbation that enhances the model's ability to process images with diverse color profiles. This is especially beneficial in human-centric vision applications where variations in lighting or color can significantly affect image perception and subsequent analysis.
\end{itemize}

\new{Region-level perturbation methods offer the advantage of simulating real-world scenarios with occlusion and information loss, contributing to a more robust and adaptable model. The simplicity of operations involved in these techniques is another strength, making them easily applicable across diverse datasets.}
\new{However, these methods exhibit certain drawbacks. While they aim to replicate real-world occlusion and information loss, there is a risk that the introduced occlusion may not authentically represent the complexity of real-world scenarios. This lack of realism in information loss poses challenges, as the model may not effectively learn to handle genuine occlusion situations. Excessive or unrealistic occlusion, stemming from overzealous removal or obscuring of image regions, can impede the model's ability to recognize crucial features. Therefore, careful management of occlusion levels is imperative to strike a balance between introducing variability and preserving informative content.}

\new{In summary, region-level perturbation methods provide a simplified means of introducing variability through simulated occlusion and information loss. Despite their operational simplicity, the challenge lies in ensuring that the introduced occlusion authentically represents real-world complexities, avoiding the risk of hindering the model's capacity to recognize essential features due to excessive or unrealistic information loss.}

\begin{figure*}[t]
  \centering
  \includegraphics[width=0.9\linewidth]{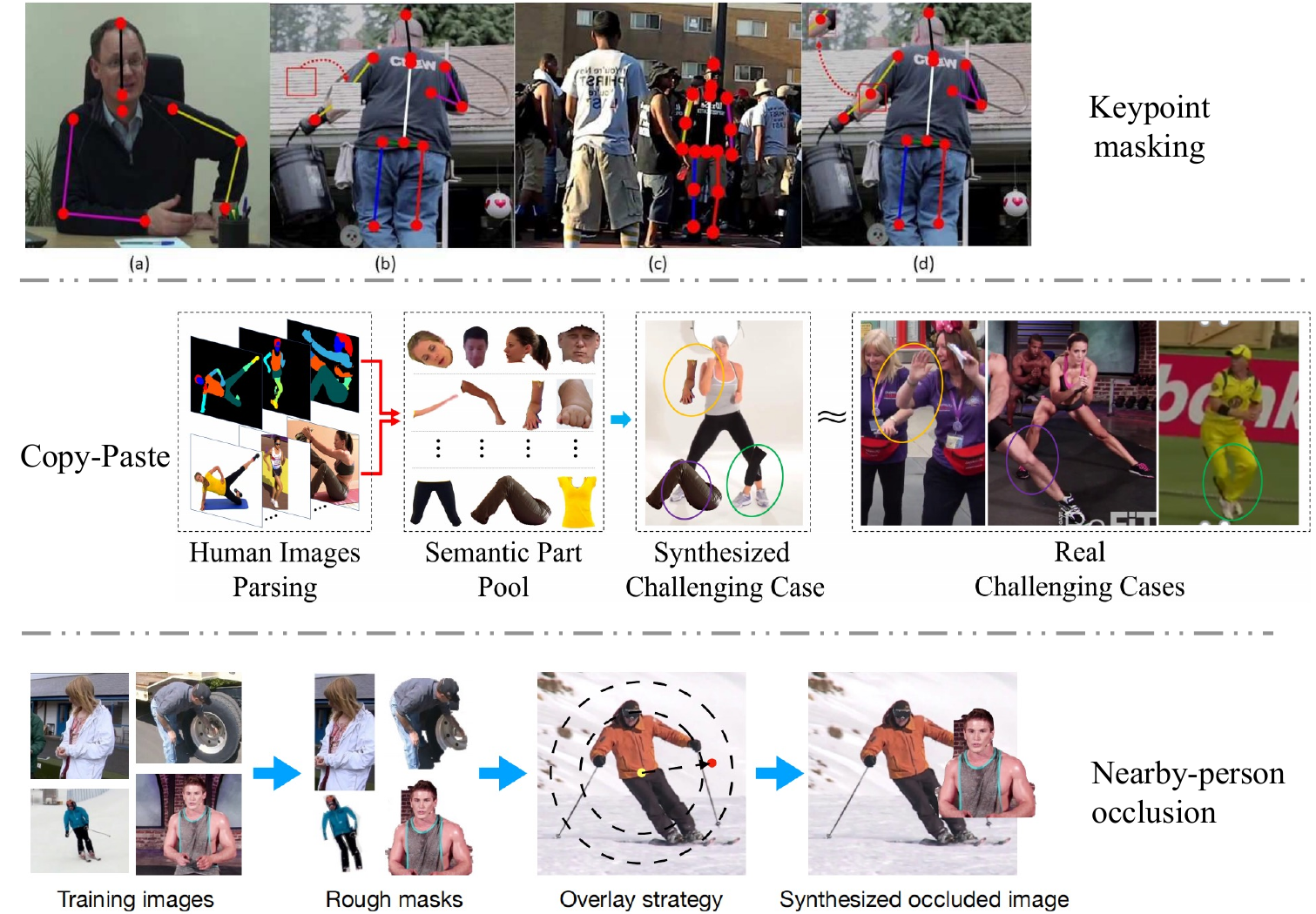}
  \caption{
    Examples of human-level occlusion generation. The figure contains representative works of Keypoint masking~\cite{ke2018multi}, Copy-Paste~\cite{bin2020adversarial} and Nearby-person occlusion~\cite{chen2021nearby}.
  }
  \label{human_occlusion}
\end{figure*}

\subsubsection{Human-level Perturbation}

The main feature of human-level data perturbation is to introduce changes at the level of a single human (instance) to increase the adaptability of the model to different instances. This may include making unique transformations to each instance to simulate the variety in the actual scene.
 
\textbf{Human-level Occlusion Generation.} 
Human-level occlusion generation, as shown in Figure~\ref{human_occlusion}, refers to the intentional introduction of obstructions or coverings over specific instances, such as human subjects, within images. This technique is applied during the training phase of machine learning models to improve their ability to handle scenarios where parts of an object or person are obscured or occluded in real-world situations.
\begin{itemize}
    \item Keypoint Masking for Occlusion Simulation~\cite{ke2018multi}: The keypoint masking technique is an innovative approach to simulate occlusion scenarios in training data. It involves two primary methods: the first method covers a body keypoint with a background patch to mimic occlusion, aiding in the model's occlusion recovery learning. The second method places body keypoint patches onto nearby background areas to simulate multiple keypoints, a scenario commonly encountered in multi-person environments. This augmentation leverages ground-truth keypoint annotations and is instrumental in training models to recognize and interpret occluded keypoints, a critical challenge in human-centric vision tasks like human pose estimation.
    \item Copy-Paste for Complex Scenario Synthesis~\cite{bin2020adversarial}: Utilizing human parsing to segment training images into various body parts, the Copy-Paste method constructs a semantic part pool based on these segments. This pool allows for the random sampling and placement of body parts onto images, synthesizing complex cases such as symmetric appearances, occlusions, and interactions with nearby individuals. By recombining body parts with different semantic granularities, this method effectively augments training data with realistic and challenging scenarios. This technique is particularly valuable in enhancing the robustness of models to complex, real-world human-centric scenarios, where occlusion and interaction between multiple persons are common.
    \item Nearby-Person Occlusion Generation~\cite{chen2021nearby}: This method focuses on creating training images that feature occlusions caused by the proximity of other individuals. Starting with a foreground human body pool generated from rough masks and keypoint annotations, this approach involves randomly sampling and placing a human body crop over another in a training image. This simulates nearby-person occlusion, a frequent occurrence in crowded or group settings. Such augmentation is crucial for training models to accurately detect and interpret human figures in dense, cluttered environments, a common challenge in applications like pedestrian detection and crowd analysis.
\end{itemize}

\new{Human-level occlusion generation methods navigate real-world scenarios featuring occluded objects or individuals.    By intentionally introducing obstructions during the training process, these techniques enrich the dataset with diverse and challenging situations, ultimately contributing to the improvement of model robustness.    The simulation of realistic occlusion scenarios enables models to develop a more nuanced understanding of object interactions and improves their adaptability to complex visual environments.}
\new{In comparison to image-level perturbation methods, human-level occlusion generation techniques provide a more nuanced and realistic simulation of occlusion scenarios, particularly in the context of human figures.    This finer granularity in simulating occlusions contributes to improved model adaptation to real-world scenarios with human occlusion, enhancing overall performance in complex visual environments.}

\new{However, the intricacies involved in handling the complex structures and relationships of human body parts may introduce increased computational overhead during the generation of occlusions.}

\begin{figure*}[t]
  \centering
  \includegraphics[width=0.9\linewidth]{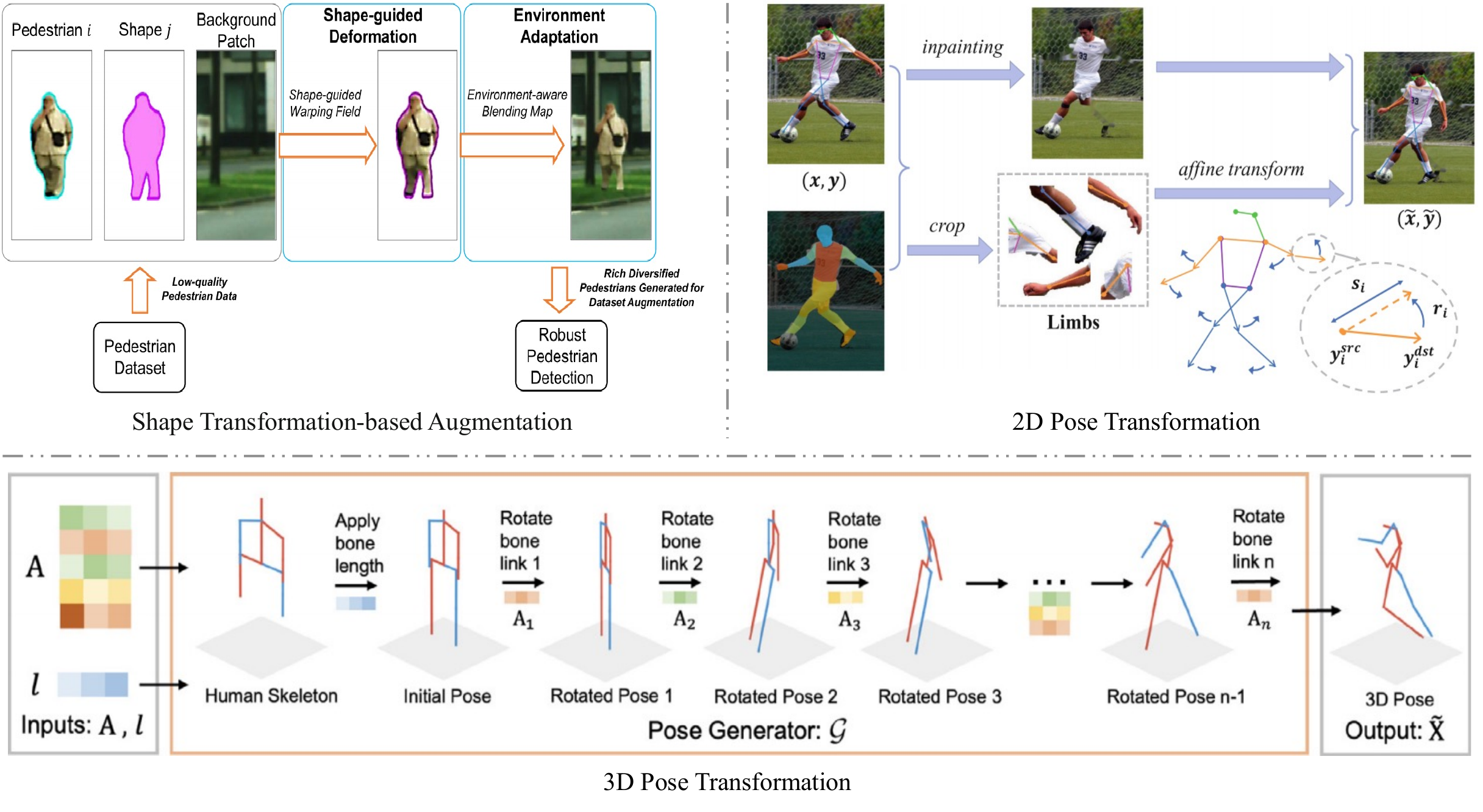}
  \caption{
    Examples of human body perturbation. The figure contains representative works of Deformable Shape Augmentation~\cite{chen2021shape}, 2D Pose Transformation~\cite{jiang2022posetrans} and 3D Pose Transformation~\cite{guan2023posegu}.
  }
  \label{human_body_per}
\end{figure*}

\textbf{Human body perturbation.}
Human body perturbation involves introducing small variations or disturbances to the positions of individual joints in a 2D/3D skeletal representation, as shown in Figure~\ref{human_body_per}. These techniques help improve the robustness and generalization capabilities of the model by exposing it to a wider range of variations in pose configurations.
\begin{itemize}
    \item Deformable Shape Augmentation~\cite{chen2021shape}: This category of data augmentation focuses on altering pedestrian shapes within images. Techniques typically involve warping fields that deform the pedestrian's shape and blending algorithms to adapt these changes to diverse environmental contexts. Such approaches are invaluable for enhancing the realism and variability of pedestrian datasets, where quality and diversity in real pedestrian images might be limited. By creating a range of shape deformations and integrating them into varying backgrounds, this method effectively prepares models for more accurate pedestrian detection and tracking in dynamically changing real-world scenarios.
    \item 2D Human Pose Transformation: Methods in this category, including approaches like PoseTrans~\cite{jiang2022posetrans}, utilize various transformation techniques, such as affine transformations, to modify the pose of 2D human figures. These methods often start with limb erasure using human parsing results, followed by selective transformation of each limb. The transformations can include scaling, rotation, and translation, generating a pool of diverse poses. Some techniques may also incorporate Generative Adversarial Networks (GANs) or pose evaluators to ensure the naturalness of generated poses. These methods are crucial for augmenting datasets in 2D human pose estimation tasks, allowing models to learn from and adapt to a wide array of human poses, improving their accuracy and robustness in real-world applications.
    \item 3D Human Pose Transformation: In the realm of 3D human pose estimation, data augmentation methods like PoseGU~\cite{guan2023posegu} focus on generating diverse 3D human poses. These techniques typically utilize skeletal models where rotation transformations and bone length adjustments are applied to generate various human poses. The goal is to create a rich dataset from minimal seed samples, significantly enhancing the diversity of poses available for training. This approach is particularly beneficial in scenarios where 3D pose data is scarce or limited to specific types of movements. By providing a broad spectrum of 3D poses, these methods greatly aid in the development of more accurate and versatile 3D human pose estimation models.
\end{itemize}

\begin{figure*}[t]
  \centering
  \includegraphics[width=0.9\linewidth]{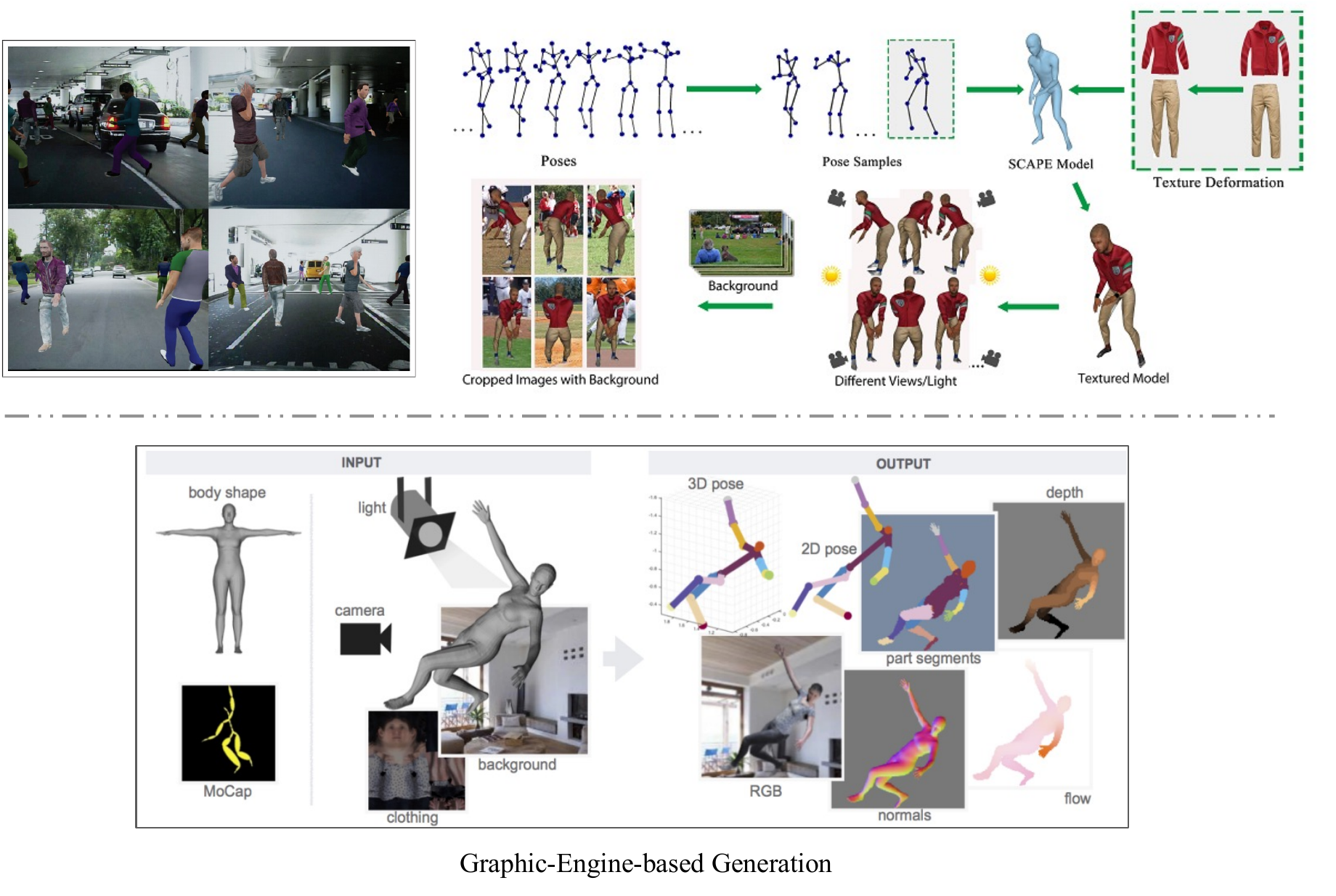}
  \caption{
    Examples of graphic engine-based generation. The figure contains representative works of MixedPeds~\cite{Cheung2017MixedPedsPD}, Synthetic Human on Real Background~\cite{chen2016synthesizing} and Synthetic Humans~\cite{varol2017learning}.
  }
  \label{graphic}
\end{figure*}

\new{Human Body Perturbation methods offer a notable advantage in their ability to intricately simulate the shapes and poses of real-world human bodies, contributing to the generation of more realistic sample data. This precision in mimicking the complexities of human forms is particularly valuable in tasks like pedestrian detection and human pose estimation, where diverse and authentic training samples are essential for model robustness.}
\new{However, these advantages come with associated drawbacks. The computational complexity linked to techniques such as Deformable Shape Augmentation, especially in the context of 3D human pose estimation, introduces a significant demand for computational resources. Consequently, the implementation of these methods may require substantial computing power and storage capacity, potentially limiting their feasibility in resource-constrained environments. Moreover, ensuring the naturalness and realism of the generated poses remains a critical consideration. In 3D human pose estimation, where the challenge lies in capturing the nuances of three-dimensional movements, the need for natural and believable poses is paramount.}

\new{In summary, while Human Body Perturbation methods excel in creating fine-grained simulations of human shapes and poses, their integration comes with computational challenges, particularly in 3D human pose estimation contexts. Striking a balance between computational demands and natural pose generation is crucial for maximizing the benefits of these methods in enhancing model robustness and generalization capabilities.}

\subsection{Data Generation}

\subsubsection{Graphic Engine-based Generation}

As shown in Figure~\ref{graphic}, synthetic data generation involves creating artificial datasets to augment training sets, employing graphics engine-based methods to produce realistic synthetic examples, and enhancing model robustness and performance.
\new{\begin{itemize}
    \item MixedPeds~\cite{Cheung2017MixedPedsPD} automatically extract the vanishing point from the dataset to calibrate the virtual camera and extract the pedestrians' scales to generate a Spawn Probability Map, which guides the algorithm to place the pedestrians at appropriate locations.  An HSV color model is used to generate clothing colors according to the color of the pedestrians. Putting synthetic human agents in the unannotated images to use these augmented images to train a Pedestrian Detector.
    \item Synthetic Human on Real Background~\cite{chen2016synthesizing} presents a fully automatic, scalable approach that samples the human pose space for guiding the synthesis procedure.  The 3D pose space is samped and the samples are used for deforming SCAPE models. Meanwhile, various clothes textures are mapped onto the human models. The deformed textured models are rendered using a variety of viewpoints and light sources and finally composited over real image backgrounds.
    \item Synthetic Humans~\cite{varol2017learning}  generates RGB images together with 2D/3D poses, surface normals, optical flow, depth images, and body-part segmentation maps for rendered people. A 3D human body model is posed using motion capture data and a frame is rendered using a background image, a texture map of the body, lighting, and a camera position.  These ingredients are randomly sampled to increase the diversity of the data. 
\end{itemize}}

Graphic engine-based generation in human-centric vision data augmentation starts with extracting key features such as pedestrian scale ratios, vanishing points, or sampling the 3D human pose space. These features guide the creation and customization of human models, often enhanced with varied clothing textures and rendered from multiple viewpoints and lighting conditions. The resulting synthetic humans are strategically placed within real-world scenes, leading to a seamless and realistic blend of virtual and real elements. Such datasets, rich in detail with features like RGB images, 2D/3D poses, and surface normals, are invaluable for a variety of applications, including pedestrian detection and human pose estimation. This approach effectively bridges the gap between simulated and natural environments, significantly enhancing model training and performance in complex, real-world scenarios.

\new{However, the use of Graphic Engine-Based Generation comes with computational demands, especially in tasks like rendering from multiple perspectives and lighting conditions. Additionally, ensuring the naturalness and authenticity of the generated scenes remains a challenge, as overly synthetic or unrealistic elements may impact the model's ability to generalize effectively.    In summary, while Graphic Engine-Based Generation enhances realism and dataset richness, careful consideration is needed to balance computational demands and maintain the authenticity required for effective model training and real-world application.}
\begin{figure*}[t]
  \centering
  \includegraphics[width=0.8\linewidth]{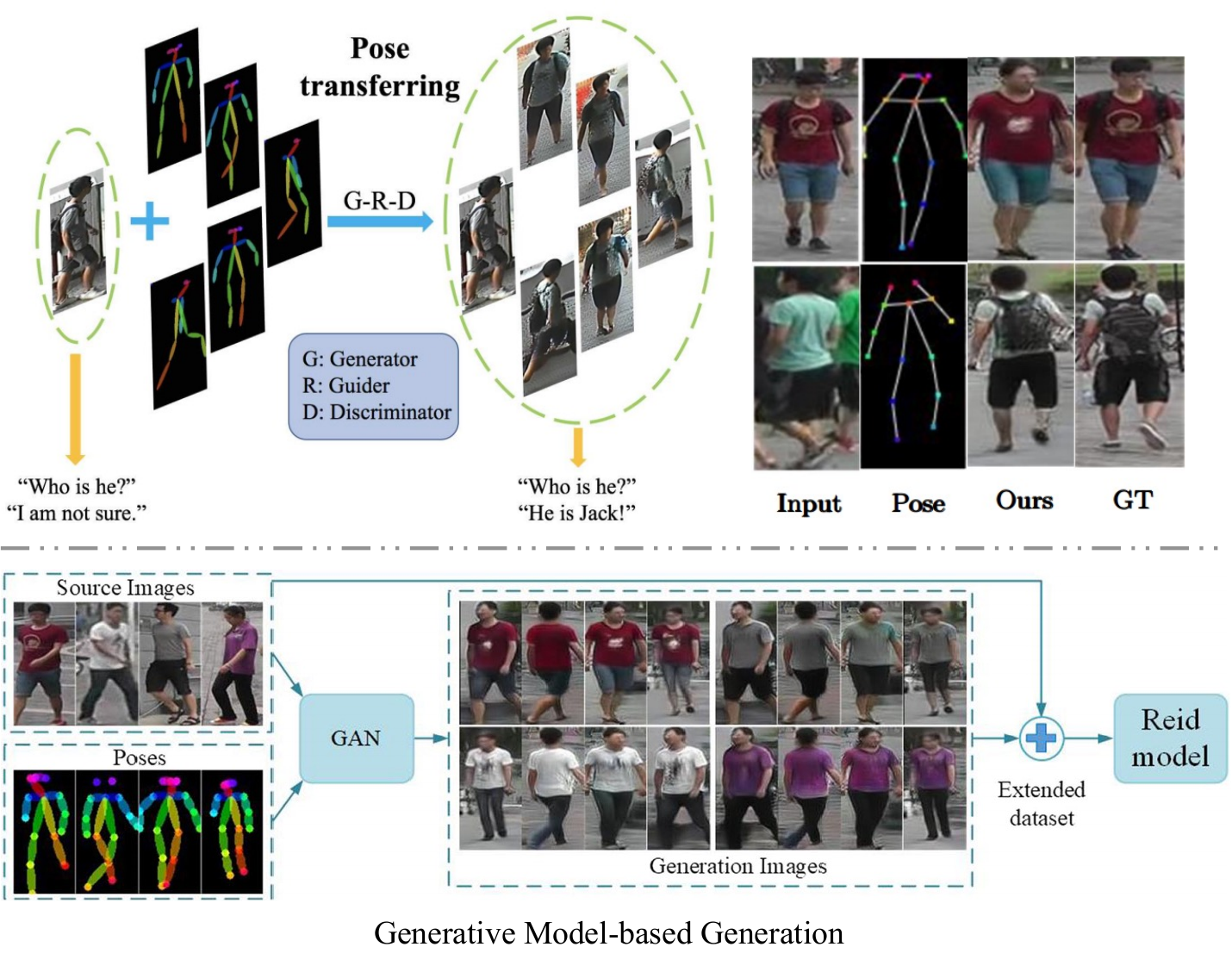}
  \caption{
    Examples of generative model-based generation. The figure contains representative works of Pose Transferring~\cite{Liu2018PoseTP}, Pose Variation Adaptation~\cite{Zhang2021PoseVA} and Pose Variation Aware Data Augmentation~\cite{Zhang2022PersonRW}
  }
  \label{generative}
\end{figure*}

\subsubsection{Generative Model-based Generation}
The generative model-based generation method, as shown in Figure~\ref{generative}, involves creating synthetic data by generating new instances or variations of existing instances within the dataset, providing the model with a richer training set.
This kind of method contributes to the model's adaptability and performance in recognizing and classifying instances under various conditions, ultimately improving its ability to handle complex and diverse datasets.
Generative Adversarial Networks are always adopted for data augmentation.
GAN-based methods~\cite{Ge2018FDGANPF,zhang2020pac,liu2020novel,yang2023improved,uc2023review} for data augmentation in human-centric vision primarily focus on enhancing the diversity and realism of training datasets through the generation of synthetic human poses. These methods leverage pose transfer GANs, often combined with modules for similarity measurement or hard example mining, to create new instances of human images in various poses. By extracting skeletal poses and pairing them with different human appearances, these techniques enable the generation of augmented data that is crucial for tasks such as person re-identification (ReID). The generated images not only enrich the pose variation in the dataset but also improve the model's ability to recognize and classify human figures across a wide range of conditions. This approach significantly contributes to the adaptability and performance of models in complex and diverse human-centric vision scenarios, making it a valuable tool in the realm of advanced data augmentation.
\new{\begin{itemize}
    \item Pose Transferring~\cite{Liu2018PoseTP} proposes a pose-transferrable person ReID framework that utilizes pose-transferred sample augmentations to enhance ReID model training.  On one hand, novel training samples with rich pose variations are generated via transferring pose instances from the MARS dataset, and they are added to the target dataset to facilitate robust training.  On the other hand, in addition to the conventional discriminator of GAN (i.e., to distinguish between REAL/FAKE samples), a novel guider sub-network that encourages the generated sample (i.e., with novel pose) towards better satisfying the ReID loss (i.e., cross-entropy ReID loss, triplet ReID loss).  
    \item Pose Variation Adaptation~\cite{Zhang2021PoseVA} proposes a pose variation adaptation method for person ReID.  It can reduce the probability of deep learning network over-fitting.  Specifically, this method introduced a pose transfer generative adversarial network with a similarity measurement module.  With the learned pose transfer model, training images can be transferred to any given pose, and with the original images, forming an augmented training dataset. 
    \item Pose Variation Aware Data Augmentation~\cite{Zhang2022PersonRW} proposes a pose transfer generative adversarial network (PTGAN).  PTGAN introduces a similarity measurement module to synthesize realistic person images that are conditional on the pose, and with the original images, form an augmented training dataset. 
\end{itemize}}

\new{Generative Adversarial Networks excel in generating realistic and diverse samples.  By employing a generator and discriminator in an adversarial training setup, GANs can produce high-quality, visually appealing outputs.  GANs are known for their ability to capture complex data distributions and generate images with intricate details.  However, GANs have certain drawbacks, including mode collapse, where the generator may focus on a limited subset of modes in the data distribution, and training instability, which can make it challenging to achieve convergence.}

\begin{figure*}[t]
  \centering
  \includegraphics[width=0.9\linewidth]{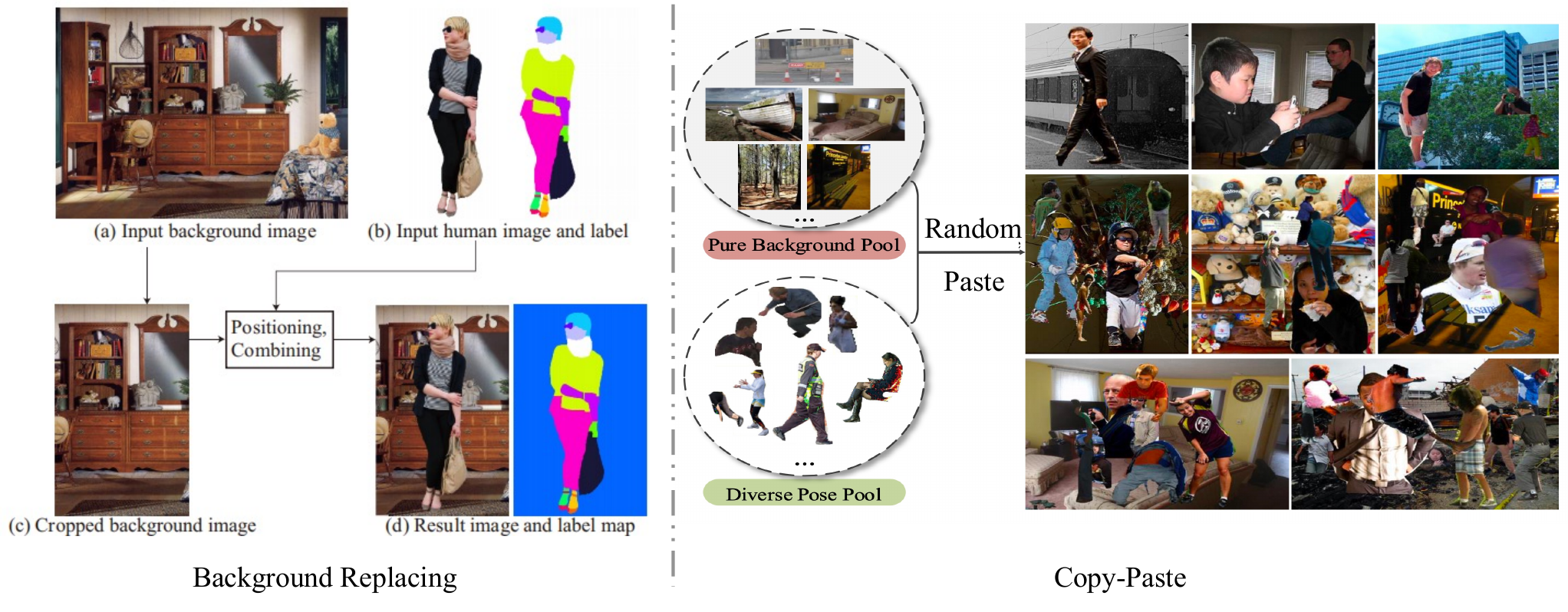}
  \caption{Examples of image-level recombination. The figure contains representative works of Background Replacing~\cite{Kikuchi2018TransferringPA} and Copy-Paste~\cite{dai2023overcoming}.
  }
  \label{image_recom}
\end{figure*}
\new{
\textbf{Diffusion Models.}
In the in-depth exploration of the future directions of diffusion models, a particularly noteworthy avenue is the application of advanced generative models, especially pre-trained Latent Diffusion Models, to enhance human-centric vision datasets. Diffusion models stand out with their distinctive paradigm for generative modeling, revolving around the iterative introduction of noise into samples to simulate the generative process of data distribution. The application of diffusion models has demonstrated significant potential in generating high-quality samples and mitigating issues associated with mode collapse. Research findings suggest that the utilization of diffusion models holds promise for achieving superior results in generating visual data, contributing to the production of more diverse and realistic images for human-centric visual tasks. Notably, the unique aspect of this approach lies in its deviation from the traditional dependence on a discriminator during the training process, potentially simplifying the entire training pipeline. This simplification could offer an effective and feasible pathway for the development and optimization of future generative models.
}
\new{In summary, within the field of generative models, the potential of latent diffusion models in guiding future research directions presents a significant outlook, introducing new possibilities for artificial intelligence applications in visual processing tasks.}

\subsubsection{Data Recombination}

\begin{figure*}[t]
  \centering
  \includegraphics[width=0.9\linewidth]{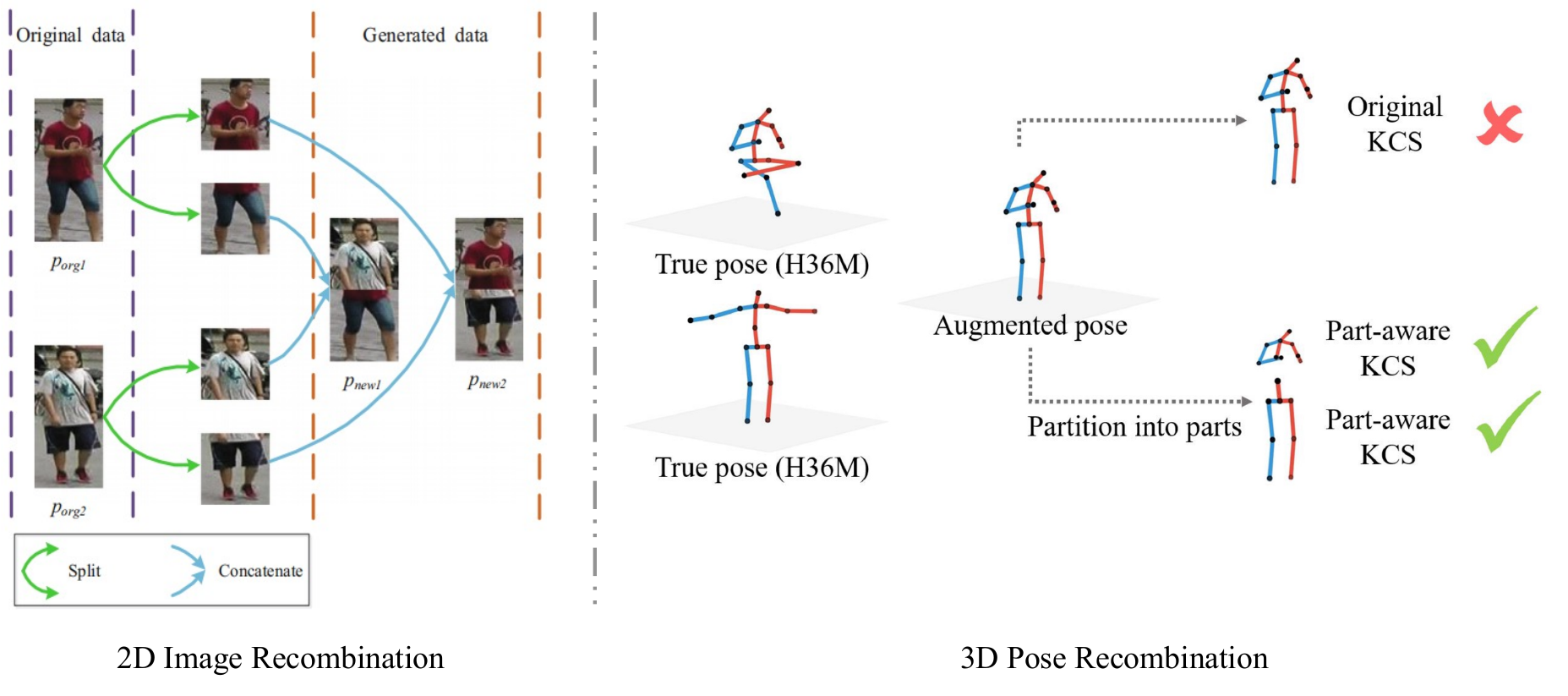}
  \caption{
    Examples of human-level recombination. The figure contains representative works of 2D Image Recombination~\cite{chen2020self} and 3D Image Recombination~\cite{gong2021poseaug}.
  }
  \label{human_recom}
\end{figure*}

\textbf{Image-level Recombination.}
Image-level recombination, as shown in Figure~\ref{image_recom}, indicates the combination and rearrangement of elements from original images, such as foreground and background recombination, background replacement, or domain mixing.
\begin{itemize}
    \item Background Replacing~\cite{McLaughlin2015DataaugmentationFR, Kikuchi2018TransferringPA,dai2023overcoming}: This method focuses on enhancing the diversity of backgrounds in human-centric images. It involves integrating a human pose estimation network to transfer features across domains and replacing original backgrounds with varied scenes from large-scale scenery datasets. Such augmentation is essential for improving model performance in environments with complex backgrounds, enhancing the robustness of human parsing networks in varied settings.
    \item Copy-paste~\cite{dwibedi2017cut,li2021cutpaste, Deng2022ImprovingCO, Remez2018LearningTS,ghiasi2021simple,fang2019instaboost}: Copy-paste augmentation represents a versatile and widely-employed technique in human-centric vision, fundamentally involving the extraction of human figures from one image and their integration into another. This method varies in its application: some approaches, like InstaBoost~\cite{fang2019instaboost}, perform copy-paste within the original image to create variations, while others construct separate pools of pure backgrounds and diverse human figures for recombination. The latter involves randomly selecting human subjects and compositing them into different backgrounds at various locations, enhancing the diversity and complexity of the training data. Some methods also take into account the spatial context and positioning, ensuring that the pasted figures align realistically with the new backgrounds. This technique is particularly effective for training models in tasks such as object detection, person re-identification, and scene understanding, as it introduces a wide range of human appearances and contextual scenarios, enriching the dataset with realistic and challenging examples.
\end{itemize}

\new{One of the primary advantages of image-level recombination is its efficiency in rapidly generating a large volume of augmented samples, thereby enriching the dataset. This process requires relatively fewer computational resources compared to certain complex augmentation methods. The ability to create diverse training samples promptly contributes to improved model robustness and generalization. Moreover, the simplicity and speed of image-level recombination make it particularly suitable for scenarios with resource constraints, enabling the augmentation of datasets at scale without excessive computational demands.}

\new{However, the generated data may exhibit a trade-off in terms of realism. The artificial nature of the recombined images might introduce a lack of authenticity, potentially leading the model to learn incorrect or unrealistic features. This limitation is crucial, especially when the goal is to train models for real-world scenarios. The challenge lies in striking a balance between the efficiency gained through rapid generation and the fidelity required for effective learning. As such, careful consideration and validation are necessary to ensure that the recombined data aligns with the actual distribution of real-world images, minimizing the risk of model learning spurious patterns that may not generalize well to authentic scenarios.}

\begin{table*}[]
\renewcommand{\arraystretch}{1.4}
\scalebox{0.75}{
\begin{tabular}{|cc|c|l|}
\hline
\multicolumn{2}{|c|}{\textbf{Tasks}} & \multicolumn{1}{c|}{\textbf{Categories}} & \multicolumn{1}{c|}{\textbf{Methods}} \\ \hline
\multicolumn{2}{|c|}{\multirow{5}{*}{\textbf{Person ReID}}} & \multirow{2}{*}{\textbf{\begin{tabular}[c]{@{}l@{}}Image-level\\ perturbation\end{tabular}}} & \begin{tabular}[c]{@{}l@{}}\textbf{Global perturbation:} (Z Zhong 2019~\cite{zhong2018camstyle})(Z Zhong 2018~\cite{zhong2018camera})\\(Z Lin 2021~\cite{lin2021color})\end{tabular} \\ \cline{4-4} 
\multicolumn{2}{|c|}{} &  & \begin{tabular}[c]{@{}l@{}}\textbf{Region-level perturbation:} (Y Gong 2021~\cite{Gong2021APR})(W Sun 2020~\cite{sun2020triplet})\end{tabular} \\ \cline{3-4} 
\multicolumn{2}{|c|}{} & \multirow{2}{*}{\textbf{\begin{tabular}[c]{@{}c@{}}Data \\ recombination\end{tabular}}} & \begin{tabular}[c]{@{}l@{}}\textbf{Image-level recombination:} (L Chen 2017~\cite{Chen2017DataGF})(N McLaughlin 2015~\cite{McLaughlin2015DataaugmentationFR})\\ (M Tian 2018~\cite{tian2018eliminating})\end{tabular} \\ \cline{4-4} 
\multicolumn{2}{|c|}{} &  & \begin{tabular}[l]{@{}l@{}}\textbf{Human-level recombination:} (K Han 2023~\cite{han2023clothing})(X Jia 2022~\cite{jia2022complementary})\\(F Chen 2020~\cite{chen2020self})\end{tabular} \\ \cline{3-4} 
\multicolumn{2}{|c|}{} & \textbf{\begin{tabular}[c]{@{}c@{}}Generative model-\\based generation\end{tabular}} & \begin{tabular}[c]{@{}l@{}}(J Liu 2018~\cite{Liu2018PoseTP})(L Zhang 2021~\cite{Zhang2021PoseVA})(L Zhang 2022~\cite{Zhang2022PersonRW})\\(D Wu 2018~\cite{wu2018random})(PAC-GAN 2020~\cite{zhang2020pac})(V Uc-Cetina 2023~\cite{uc2023review})\\(Z Yang 2023~\cite{yang2023improved})(Q Wu 2021~\cite{wu2021deep})\end{tabular} \\ \hline
\multicolumn{1}{|c|}{\multirow{8}{*}{\textbf{\begin{tabular}[c]{@{}c@{}}Human Pose\\ Estimation\end{tabular}}}} & \multirow{5}{*}{\textbf{2D}} & \multirow{2}{*}{\textbf{\begin{tabular}[c]{@{}c@{}}Image-level\\ perturbation\end{tabular}}} & \textbf{Global perturbation:} (X Peng 2018~\cite{peng2018jointly})(Wang 2021~\cite{wang2021human}) \\ \cline{4-4} 
\multicolumn{1}{|c|}{} &  &  & \textbf{Region-level perturbation:} (J Huang 2020~\cite{huang2020aid}) \\ \cline{3-4} 
\multicolumn{1}{|c|}{} &  & \textbf{\begin{tabular}[c]{@{}c@{}}Data\\ recombination\end{tabular}} & \textbf{Image-level recombination:} (Instaboost 2019~\cite{fang2019instaboost})(Dai 2022~\cite{dai2023overcoming}) \\ \cline{3-4} 
\multicolumn{1}{|c|}{} &  & \multirow{2}{*}{\textbf{\begin{tabular}[c]{@{}c@{}}Human-level\\ perturbation\end{tabular}}} & \begin{tabular}[c]{@{}l@{}}\textbf{Human-level occlusion generation:} (L Ke 2018~\cite{ke2018multi})\\(Y Bin 2020~\cite{bin2020adversarial})(Y Chen 2021~\cite{chen2021nearby})\end{tabular} \\ \cline{4-4} 
\multicolumn{1}{|c|}{} &  &  & \textbf{Human body perturbation:} (W Jiang 2022~\cite{jiang2022posetrans}) \\ \cline{2-4} 
\multicolumn{1}{|c|}{} & \multirow{3}{*}{\textbf{3D}} & \textbf{\begin{tabular}[c]{@{}c@{}}Human-level\\ perturbation\end{tabular}} & \begin{tabular}[c]{@{}l@{}}\textbf{Human body perturbation: }(Li 2021~\cite{li2020cascaded})(Z Xin 2022~\cite{xin20223d})\\ (L Huang 2022~\cite{huang2022dh})(PoseGU 2023~\cite{guan2023posegu})\end{tabular} \\ \cline{3-4} 
\multicolumn{1}{|c|}{} &  & \textbf{\begin{tabular}[c]{@{}c@{}}Data\\ recombination\end{tabular}} & \textbf{Human-level recombination: }(Gong 2022~\cite{gong2021poseaug}) \\ \cline{3-4} 
\multicolumn{1}{|c|}{} &  & \textbf{\begin{tabular}[c]{@{}c@{}}Graphic engine-\\based generation\end{tabular}} & (Rogez 2016~\cite{rogez2016mocap})(Mehta 2017~\cite{mehta2017vnect})(W Chen 2017~\cite{chen2016synthesizing})(Varol 2018~\cite{varol2017learning}) \\ \hline
\multicolumn{2}{|c|}{\textbf{Human Parsing}} & \textbf{\begin{tabular}[c]{@{}c@{}}Data\\ recombination\end{tabular}} & \begin{tabular}[c]{@{}l@{}}\textbf{Image-level recombination:} (T Kikuchi 2017~\cite{Kikuchi2018TransferringPA})(T Remez 2018~\cite{Remez2018LearningTS})\\ (G Ghiasi 2020~\cite{ghiasi2021simple})(Instaboost 2019~\cite{fang2019instaboost})\end{tabular} \\ \hline
\multicolumn{2}{|c|}{\multirow{6}{*}{\textbf{\begin{tabular}[c]{@{}l@{}}Pedestrian\\ Detection\end{tabular}}}} & \multirow{2}{*}{\textbf{\begin{tabular}[c]{@{}c@{}}Image-level\\ perturbation\end{tabular}}} & \begin{tabular}[c]{@{}l@{}}\textbf{Region-level perturbation:} (Z Zhong 2020~\cite{zhong2020random})(S Cygert 2020~\cite{Cygert2020TowardRP})\\(Pedhunter 2020~\cite{chi2020pedhunter})\end{tabular} \\ \cline{4-4} 
\multicolumn{2}{|c|}{} &  & \textbf{Global perturbation:} (C Michaelis 2019~\cite{michaelis2019benchmarking}) \\ \cline{3-4} 
\multicolumn{2}{|c|}{} & \textbf{\begin{tabular}[c]{@{}c@{}}Data\\ recombination\end{tabular}} & \begin{tabular}[c]{@{}l@{}}\textbf{Image-level recombination:} (D Dwibedi 2017~\cite{dwibedi2017cut})(CL Li 2021~\cite{li2021cutpaste})\\(J Deng 2022~\cite{Deng2022ImprovingCO})\end{tabular} \\ \cline{3-4} 
\multicolumn{2}{|c|}{} & \textbf{\begin{tabular}[c]{@{}c@{}}Human-level\\ perturbation\end{tabular}} & \textbf{Human body perturbation:} (Z Chen 2019~\cite{chen2021shape}) \\ \cline{3-4} 
\multicolumn{2}{|c|}{} & \textbf{\begin{tabular}[c]{@{}c@{}}Graphic engine-\\based generation\end{tabular}} & (J Nilsson 2014~\cite{nilsson2014pedestrian})(MixedPeds 2017~\cite{Cheung2017MixedPedsPD})(SynPoses 2022~\cite{nie2022synposes}) \\ \cline{3-4} 
\multicolumn{2}{|c|}{} & \textbf{\begin{tabular}[c]{@{}c@{}}Generative model-\\based generation\end{tabular}} & (Bo Lu 2022~\cite{Lu2022PedestrianDF})(R Zhi 2021~\cite{Zhi2021PoseGuidedPI})(X Zhang 2020~\cite{zhang2020deep})(S Liu 2020~\cite{liu2020novel}) \\ \hline
\end{tabular}
}
\vspace{2mm}
\caption{Categorized by human-centric vision tasks.}
\label{tab:category2}
\end{table*}

\textbf{Human-level Recombination.}
The human-level recombination method, as shown in Figure~\ref{human_recom}, involves manipulating image data by selectively extracting and relocating human parts within an image or across images to generate synthetic examples.
\begin{itemize}
    \item 2D Human Image Recombination: This technique~\cite{chen2020self} involves splitting 2D human images into distinct upper and lower parts and then creatively recombining these segments to generate new synthetic examples. The method effectively increases the diversity of human appearances and postures within the training dataset, introducing variations that are essential for robust model training. By reassembling different combinations of human features, this augmentation strategy enhances the ability of models to accurately recognize and classify individuals or detect pedestrians in varied scenarios. 
    \item 3D Human Pose Recombination: PoseAug~\cite{gong2021poseaug} exemplifies a human-level manipulation approach in data augmentation, focusing on enhancing pose diversity for 3D human pose estimation. It introduces a novel pose augmentor capable of adjusting various geometric factors such as posture, body size, viewpoint, and position through differentiable operations. This capability allows for the augmentor to be jointly optimized with the 3D human pose estimator, using estimation errors as feedback to generate more diverse and challenging poses in an online manner. A key feature of PoseAug is its ability to split and recombine the upper and lower parts of the human body in 3D poses, creating new, synthetic examples. 
\end{itemize}

\new{One notable advantage of human-level recombination lies in its capability to introduce nuanced and realistic variations into the dataset.   By selectively extracting and relocating human parts, this method can simulate a wide range of scenarios, such as occlusions, interactions, and variations in body poses.   This diversity contributes to the creation of a more comprehensive training dataset, enhancing the model's adaptability to complex, real-world conditions.}

\new{On the other side, the realism of the generated synthetic examples may be subject to the precision and appropriateness of the recombination process. Inaccuracies in part extraction or improper relocation could result in unrealistic images that deviate from the natural distribution of real-world data.   This challenge poses a risk of the model learning patterns or features that may not be representative of authentic scenarios.   Consequently, careful validation and refinement of the human-level recombination process are essential to ensure that the generated samples maintain authenticity and contribute positively to the model's robustness.   Balancing the introduction of diversity with the preservation of realism is crucial for maximizing the effectiveness of human-level recombination in data augmentation.}

\section{Categorized by Human-Centric Vision Tasks}
\label{sec3}


In Table~\ref{tab:category2}, we classify data augmentation in human-centric vision into two main branches: data perturbation and data augmentations.
The former indicates methods that perturb the original example for data augmentation, while the latter refers to methods that generate training new examples for data augmentation.
The specifics of each data augmentation method are thoroughly discussed in subsequent sections.

\subsection{Person ReID}

Person Re-Identification (ReID) in computer vision is a challenging task that involves recognizing and matching individuals across different camera views. This task becomes crucial in surveillance and security applications, where the goal is to track individuals' movements without compromising personal privacy. Advanced ReID systems utilize deep learning techniques to analyze features like clothing, gait, and even subtle physical characteristics, striving to achieve high accuracy even in crowded or dynamic environments. The main challenge in ReID lies in handling variations in lighting, pose, and occlusion, making robust feature extraction and matching essential for effective identification.

\begin{table}
    \centering
    \begin{tabular}{|c|c|c|} \hline
        Methods &  mAP (↑)&  Rank-1 (↑)\\ \hline
        \multicolumn{3}{|c|}{Baselines}\\ \hline
        SVDNet~\cite{Sun2017SVDNetFP}& 62.1 & 82.3\\ \hline
        \multicolumn{3}{|c|}{Data augmentation methods}\\ \hline
        DeformGAN~\cite{siarohin2018deformable} & 61.3 &80.6 \\
        LRSO~\cite{zheng2017unlabeled} &66.1 &84.0 \\
        Random erasing~\cite{zhong2020random}&71.3& 87.0\\
        CamStyle~\cite{zhong2018camera} &71.6 &89.5 \\
        PN-GAN~\cite{qian2018pose} &72.6 &89.4\\
        FD-GAN~\cite{ge2018fd} &77.7 &90.5\\
        DG-Net~\cite{zheng2019joint} &86.0 &94.8\\ \hline
    \end{tabular}
    \caption{Person Re-identification Performance Comparison of Methods with Data Augmentation on Market1501 Dataset. We compared data augmentation methods based on SVDNet with ResNet-50 as the backbone.}
    \label{tab:reid}
\end{table}

\begin{figure*}[t]
  \centering
  \includegraphics[width=0.8\linewidth]{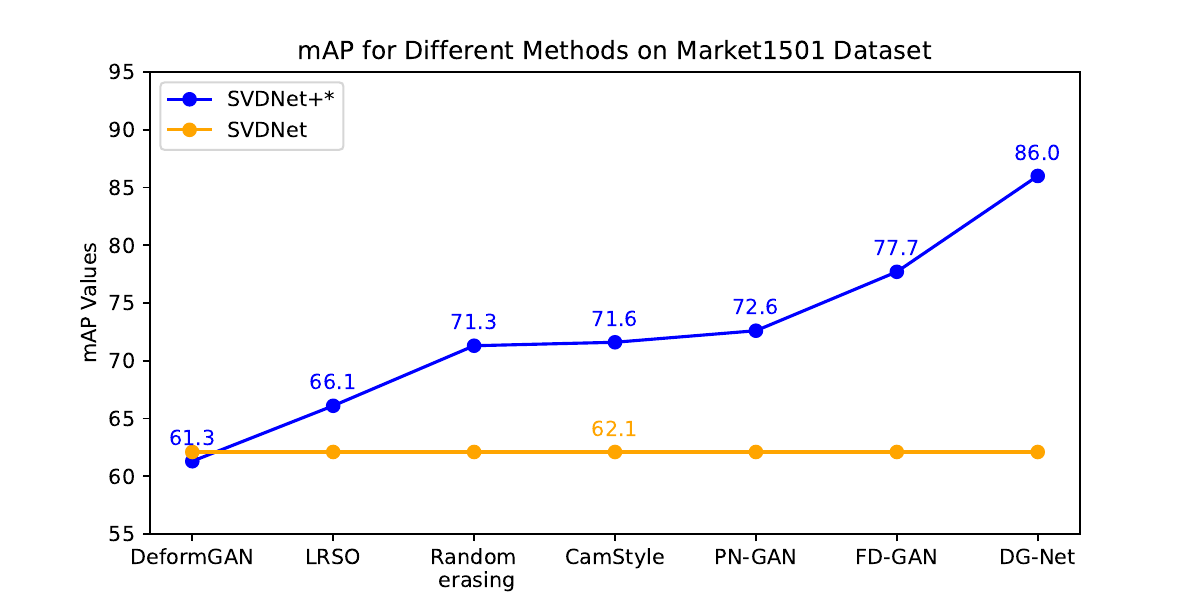}
  \caption{
\new{mAP of different methods on Market1501 Dataset. The orange line represents the baseline SVDNet, while the blue line represents the baseline combined with various data augmentation methods.  }}
  \label{reid1}
\end{figure*}
\begin{figure*}[t]
  \centering
  \includegraphics[width=0.8\linewidth]{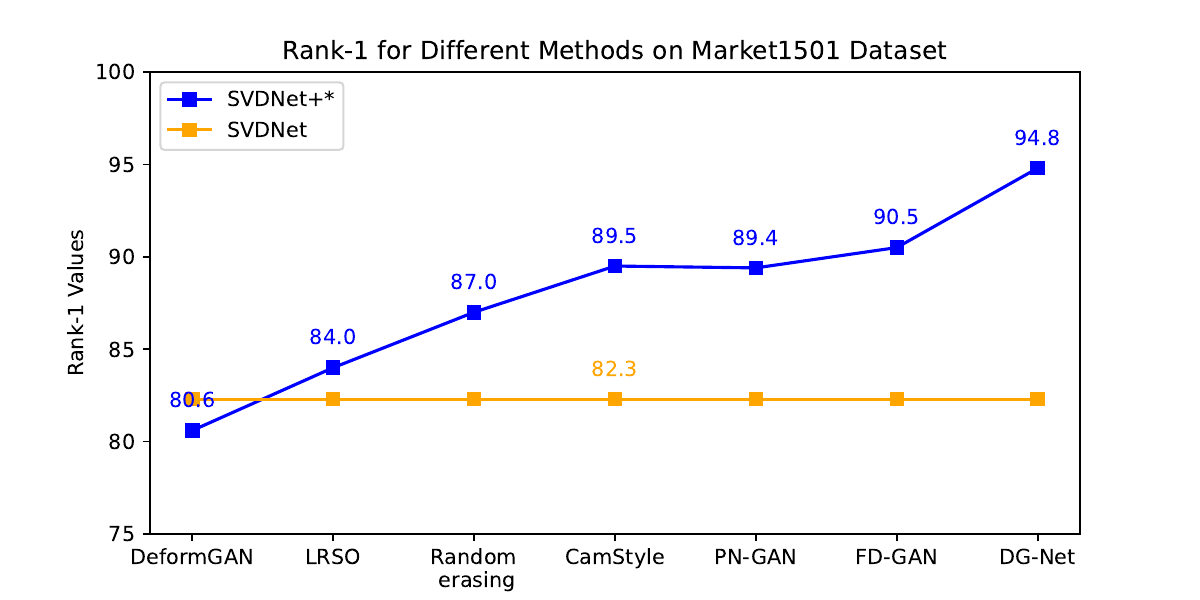}
  \caption{
\new{Most of the data augmentation methods outperformed the baseline in the rank-1 metric. The data augmentation method of generating a human body image is more effective than the image-level perturbation.}  }
  \label{reid2}
\end{figure*}

The data augmentation methods in person re-identification (ReID) tasks include image-level perturbation, image/human-level recombination, and generative model-based data generation.
Image-level perturbation techniques, such as camera style transfer~\cite{zhong2018camstyle,zhong2018camera} and grayscale patch augmentation~\cite{Gong2021APR}, enable ReID models to adapt to different camera styles and environmental conditions by altering color information and encouraging them to focus on structural features. Data recombination augmentation involves background replacing~\cite{McLaughlin2015DataaugmentationFR,chen2016synthesizing,tian2018eliminating} at the image level and human instance recombination at the human level~\cite{chen2020self}, which generates new backgrounds and human appearances, respectively, increasing the diversity of training data.  Generative model-based data generation methods employ GANs~\cite{wu2019deep, zhang2020pac, uc2023review, yang2023improved, wu2021deep, siarohin2018deformable} to synthesize human images with varied clothing and textures while maintaining consistent identity.  This significantly enriches the dataset for each identity and improves the ReID model's ability to generalize across different appearances and poses.  These augmentation strategies collectively enhance the robustness and generalizability of ReID systems, enabling them to perform effectively in diverse surveillance and security contexts.




\new{\textbf{Dataset.}
In the context of Person Re-identification (ReID) tasks, prominent datasets such as Market-1501~\cite{zheng2015scalable}, DukeMTMC-reID~\cite{ristani2016performance}, and CUHK03~\cite{li2014deepreid} are commonly employed. In this study, our primary focus lies on the evaluation of the efficacy of diverse data augmentation methods using the Market-1501 dataset. Market-1501 furnishes an environment simulating real-world surveillance scenarios, thereby endowing algorithms with enhanced robustness for practical applications. The dataset, originating from multiple surveillance cameras, encompasses over 1500 identities and a total of 32,000 images. Notably, it incorporates challenging factors such as disparate backgrounds, varied poses, and occlusions, rendering it an ideal benchmark for assessing the performance of Person ReID algorithms. The chosen dataset, with its realistic surveillance setting and diverse challenges, serves as a pertinent backdrop for evaluating the impact of augmentation strategies on the robustness and generalization capabilities of Person ReID models.}

\textbf{Person re-identification experimental analysis.}
We collect and compare the performance of various methods with data augmentation for person re-identification on the Market1501 dataset using mAP (mean Average Precision) and Rank-1 accuracy as evaluation metrics in Table~\ref{tab:reid}. \new{The data augmentation methods show good performance, as shown in Figure~\ref{reid1} and~\ref{reid2}. All methods use SVDNet as the baseline and ResNet-50 as the backbone.}
To assess the effectiveness of data augmentation methods, we employed SVDNet~\cite{Sun2017SVDNetFP} as the baseline. In the data augmentation methods, based on the evaluation metrics, DG-Net~\cite{zheng2019joint} demonstrates the highest performance among all the methods, followed by FD-GAN~\cite{ge2018fd}, PN-GAN~\cite{qian2018pose}, and CamStyle~\cite{zhong2018camstyle}. These methods exhibit better performance compared to the other data augmentation methods and the baselines. The results showcase the effectiveness of data augmentation for improving the performance of human-centric vision tasks such as person re-ID.

\subsection{Human Pose Estimation}

Human pose estimation involves detecting the positions and orientations of human joints in images or videos, aiming to understand human body language and actions. This technology has far-reaching implications, particularly in sports analytics, physical therapy, and entertainment. By accurately tracking joint positions, human pose estimation algorithms enable the analysis of body movements, offering feedback for performance improvement in athletes or monitoring rehabilitation progress in patients. The major challenge in human pose estimation is achieving accurate joint detection in real-time, especially in complex environments where occlusions or rapid movements occur.

\subsubsection{2D Human Pose Estimation}

For 2D human pose estimation tasks, the data augmentation methods focus on image-level perturbation and human-level perturbation.
Image-level perturbation involves direct modifications to the image. Techniques such as learnable scaling and rotating transformations, as applied in Peng~\cite{peng2018jointly}, adjust the orientation and size of images to simulate different viewing angles and distances. Wang~\cite{wang2021human} proposes the injection of noise into images, enhancing the model's robustness against variations in image quality and real-world disturbances. These perturbations at the image level ensure that the human pose estimation models are not only accurate but also adaptable to a wide range of imaging conditions.

Human-level perturbation introduces more variance specific to humans. Methods like those presented in Ke~\cite{ke2018multi} and Chen~\cite{chen2021nearby} create occlusions in keypoint areas, simulating real-world scenarios where humans are partially obscured by objects or other people. Another innovative approach involves cutting and pasting human limbs from different images (as shown in Bin~\cite{bin2020adversarial}), replicating scenarios of overlapping and interacting human figures. These occlusions and limb manipulations closely mimic complex, crowded environments, providing a more comprehensive training ground for human pose estimation models. An advanced human body perturbation method, as described in Jiang~\cite{jiang2022posetrans}, involves the direct transformation of limbs in the original image. This technique modifies the position, rotation, and size of specific body parts, offering a direct and effective means of simulating a wide range of human poses and movements. Such direct manipulation of the human figure itself presents a unique challenge for human pose estimation models, pushing them to accurately detect joints and limb orientations even under substantial alterations.

\begin{table*}
    \centering
    \begin{tabular}{|c|c|c|c|c|} \hline  
         Method&  AP (↑)&  AP50 (↑)&  AP75 (↑)& AR (↑)\\ \hline  
        \multicolumn{5}{|c|}{Baseline}\\ \hline  
         HRNet-W32~\cite{toshev2014deeppose}&  74.4&  90.5&  81.9& 79.8\\ \hline  
        \multicolumn{5}{|c|}{Data augmentation methods}\\ \hline  
         +Cutout*~\cite{devries2017improved}&  74.5&  90.5&  81.7& 78.8\\
         +GridMask~\cite{chen2020gridmask}&  74.7&  90.6&  82.0& 80.1\\
         +Photometric Distortion~\cite{bochkovskiy2020yolov4}&  74.6&  90.3&  81.9& 80.0\\
         +AdvMix~\cite{wang2021human}&  74.7&  -&  -& -\\
         +InstaBoost~\cite{fang2019instaboost}&  74.7&  90.5&  82.0& 80.1\\  
         +ASDA~\cite{bin2020adversarial}&  75.2&  91.0&  82.4& 80.4\\  
         +PoseTrans~\cite{jiang2022posetrans}&  75.5&  91.0&  82.9& 80.7\\ \hline
    \end{tabular}
    \caption{2D human pose estimation performance comparison of methods with data augmentation on MS-COCO val set. Results marked with ‘*’ are using CascadeRCNN bounding boxes. We compared data augmentation methods based on the HRNet-W32 backbone.}
    \label{tab:2dpose}
\end{table*}

\begin{figure*}[t]
  \centering
  \includegraphics[width=0.8\linewidth]{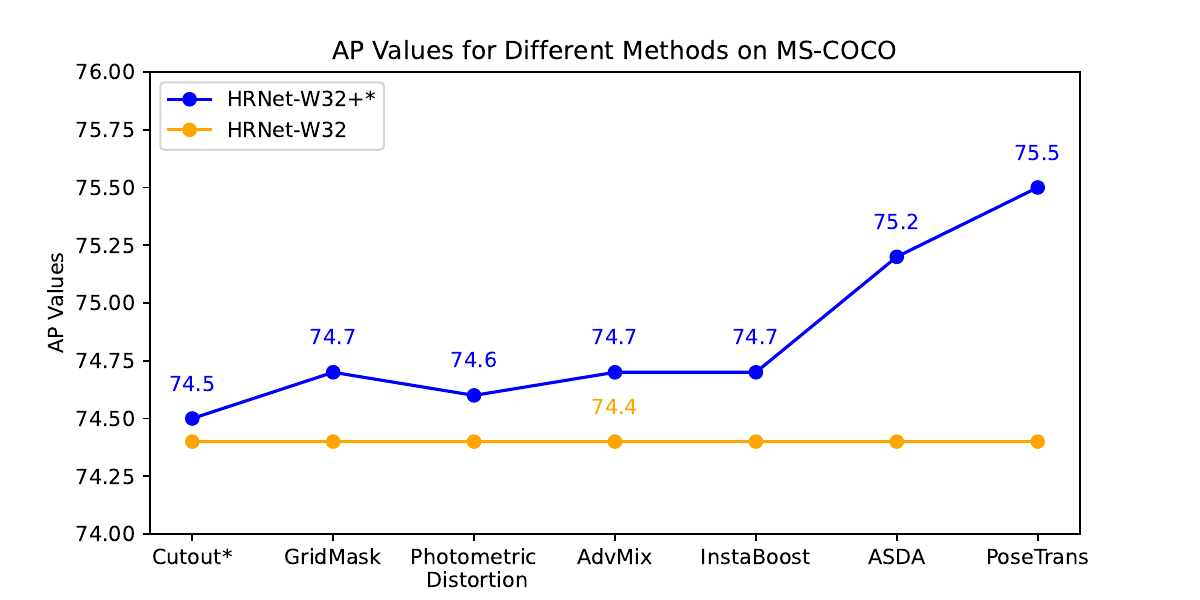}
  \caption{
\new{AP of different methods on MS-COCO Dataset. The orange line represents the baseline HRNet-W32, while the blue line represents the baseline combined with various data augmentation methods. Human-level perturbation methods like PoseTrans can usually get better results.} }
  \label{2dpose}
\end{figure*}

\begin{figure*}[t]
  \centering
  \includegraphics[width=1.0\linewidth]{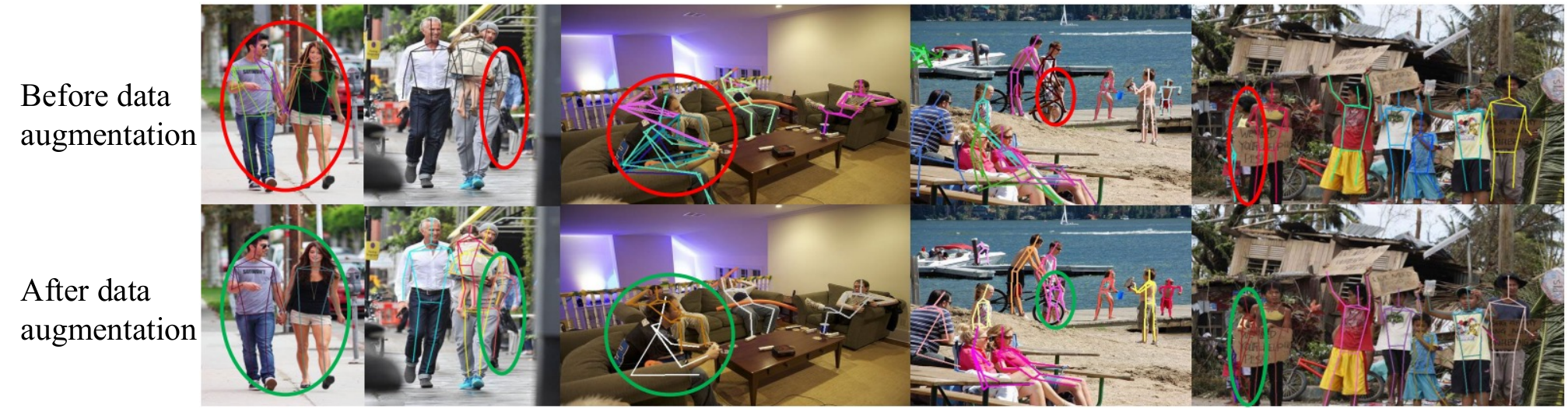}
  \caption{
\new{Comparison of data augmentation method and baseline performance on the CrowdHuman Dataset~\cite{shao2018crowdhuman}. Red circles emphasize the inaccurate keypoints predicted by HigherHRNet~\cite{cheng2020higherhrnet}, while green circles demonstrate predictions of data augmentation method Full-DG~\cite{dai2023overcoming}.}  }
  \label{2dpose_vis}
\end{figure*}

\new{\textbf{Dataset.}
For the task of 2D human pose estimation, commonly utilized datasets include MS-COCO~\cite{lin2014microsoft} and MPII Human Pose Dataset~\cite{andriluka20142d}. In the ensuing discussion, we aim to assess the effectiveness of various data augmentation methods specifically in the realm of 2D human pose estimation, with a primary focus on the MS-COCO dataset. The MS-COCO dataset comprises over 200,000 images, spanning 80 different object categories, encompassing entities such as humans, animals, vehicles, furniture, and more. With detailed annotations for the task of human pose estimation, each human instance is meticulously labeled, associating keypoint markers. The MS-COCO dataset holds significant value for advancing research in 2D human pose estimation algorithms.}

\textbf{2D human pose estimation experimental analysis.}
Table~\ref{tab:2dpose} shows the comparison of the 2D human pose estimation performance of different methods on the MS-COCO validation set. The evaluation metrics used are AP (average precision), AP50, AP75, and AR (average recall). \new{All methods are compared against the same baseline using HRNet-W32 as the backbone. Various Baseline-based data augmentation methods have seen improvements, as shown in Figure~\ref{2dpose}. Data-augmented models tend to improve in various ways, such as better detecting human pose in dense crowds, as shown in Figure~\ref{2dpose_vis}.}
The baseline method, HRNet-W32~\cite{toshev2014deeppose}, achieves a moderate level of performance across the evaluation metrics. The data augmentation methods, including Cutout~\cite{devries2017improved}, GridMask~\cite{chen2020gridmask}, Photometric Distortion~\cite{bochkovskiy2020yolov4}, AdvMix~\cite{wang2021human}, InstaBoost~\cite{fang2019instaboost}, ASDA~\cite{bin2020adversarial}, and PoseTrans~\cite{jiang2022posetrans}, show improvements in performance compared to the baseline method. It appears that some data augmentation methods, such as PoseTrans~\cite{jiang2022posetrans}, demonstrate higher performance than others, as indicated by higher AP, AP50, AP75, and AR scores. Overall, Table~\ref{tab:2dpose} suggests that employing data augmentation techniques can enhance the performance of 2D human pose estimation models on the MS-COCO dataset. Experimenting with different augmentation methods, such as PoseTrans, may yield better results in terms of accuracy.

\subsubsection{3D Human Pose Estimation}

In the field of 3D human pose estimation, data augmentation techniques involve various methods, including those generated in graphics engines such as graphic engine-based generation, and human-level perturbation recombination.
Compared with 2D human pose estimation, 3D pose ground truth is far more difficult to obtain. Most of the existing 3D human pose estimation datasets are collected indoors, which results in poor generalization in real-world applications.
Thus, data augmentation is a very important technique for 3D human pose estimation.

\begin{table}
    \centering
    \begin{tabular}{|c|c|} \hline
        Method &MPJPE (↓)\\ \hline
 \multicolumn{2}{|c|}{Non-data augmentation methods}\\ \hline
        SemGCN~\cite{zhao2019semantic} & 57.60\\
        Sharma~\cite{sharma2019monocular}  & 58.00\\
        Moon~\cite{moon2019camera}  & 54.40\\
        VPose~\cite{pavllo20193d}& 52.70 \\ \hline
 \multicolumn{2}{|c|}{Data augmentation methods}\\ \hline
        Li~\cite{li2020cascaded}  & 50.90\\
        VPose + PoseAug~\cite{gong2021poseaug} & 50.20\\
        VPose + DH-AUG~\cite{huang2022dh} & 49.81 \\ \hline
    \end{tabular}
    \caption{3D human pose estimation performance comparison of methods with Data Augmentation on H36M dataset. For 3D human pose estimation tasks, there is no standardized backbone for data augmentation methods.}
    \label{tab:3dpose1}
    \vspace{-6mm}
\end{table}

\begin{figure*}[t]
  \centering
  \includegraphics[width=0.8\linewidth]{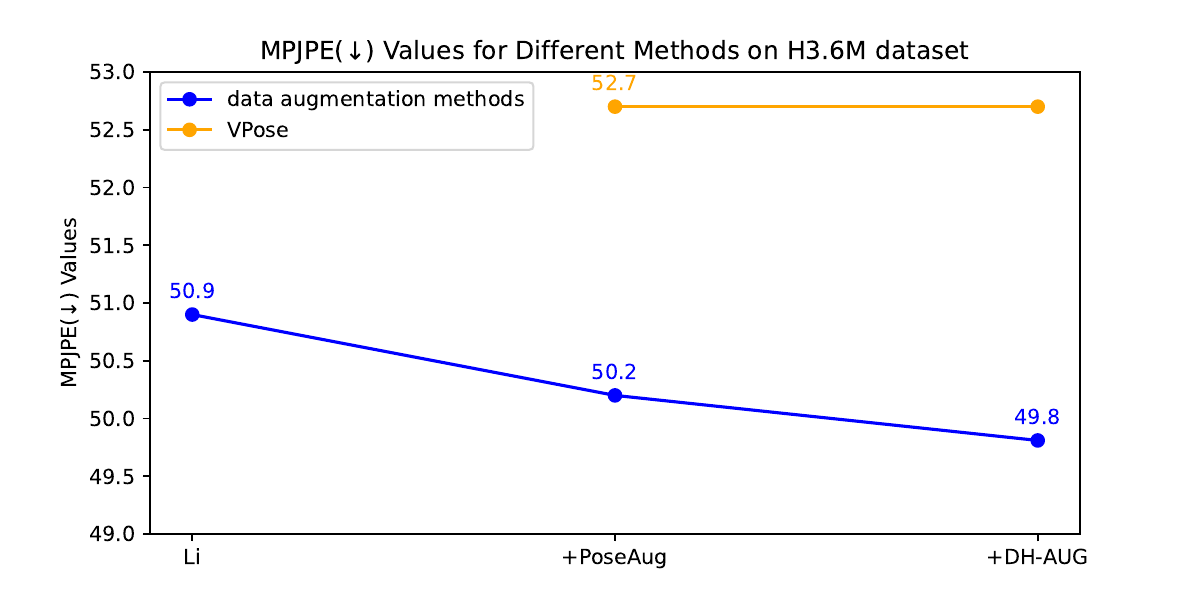}
  \caption{
\new{MPJPE of different methods on H3.6M Dataset. Compared to the baseline, the results are significantly better after using the data augmentation methods.}  }
  \label{3dpose1}
\end{figure*}

Graphic engine-based generation is a specific data augmentation strategy for obtaining 3D ground truth. This kind of method employs simulator or rendering methods to synthesize 3D human instances and paste them into the diverse real background. It helps create training examples with almost zero cost and greatly increases the number of training set. 
Generative model-based generation is also adopted, Rogez et al.~\cite{rogez2016mocap} introduce an image-based synthesis engine that artificially augments a dataset of real images with 2D human pose annotations using 3D motion capture data.

Apart from generation, human-level perturbation aims to perform the transformation in the limb of 3D pose by rotating and scaling. Human-level recombination for 3D human pose estimation aims to split the 3D human pose into upper/lower parts and then recombine them.

\begin{table*}
    \centering
    \begin{tabular}{|c|c|c|c|c|} \hline
        Methods &  CE &MPJPE (↓) &PCK (↑) &AUC (↑)\\ \hline
 \multicolumn{5}{|c|}{Non-data augmentation methods}\\ \hline
        Multi Person~\cite{chu2017multi} & & 122.20 &75.20 &37.80 \\
        OriNet~\cite{luo2018orinet} & &89.40 &81.80 &45.20 \\
        LCN~\cite{ci2019optimizing} & \checkmark & - &74.00 &36.70 \\
        HMR~\cite{kanazawa2018end} & \checkmark &113.20 &77.10 &40.70\\
        SRNet~\cite{zeng2020srnet} & \checkmark &- &77.60 &43.80 \\
        RepNet~\cite{wandt2019repnet} & \checkmark &92.50 &81.80 &54.80\\ 
        VPose~\cite{pavllo20193d}& \checkmark & 86.60 &- &- \\ \hline
 \multicolumn{5}{|c|}{Data augmentation methods}\\ \hline
        VNect~\cite{mehta2017vnect} & & 124.70 &76.60 &40.40 \\
        Mehta~\cite{mehta2017monocular}& &117.60 &76.50 &40.80 \\
        Li~\cite{li2020cascaded} & \checkmark &99.70 &81.20 &46.10 \\
        VPose+PoseAug~\cite{gong2021poseaug} & \checkmark &73.00 &88.60 &57.30 \\
        VPose+DH-AUG~\cite{huang2022dh} & \checkmark &71.17 &89.45 &57.93 \\ \hline
    \end{tabular}
    \caption{3D human pose estimation performance comparison of methods with Data Augmentation on 3DHP dataset. CE means evaluation across datasets. For 3D human pose estimation tasks, there is no standardized backbone for data augmentation methods.}
    \label{tab:3dpose2} 
\end{table*}

\begin{figure*}[t]
  \centering
  \includegraphics[width=0.8\linewidth]{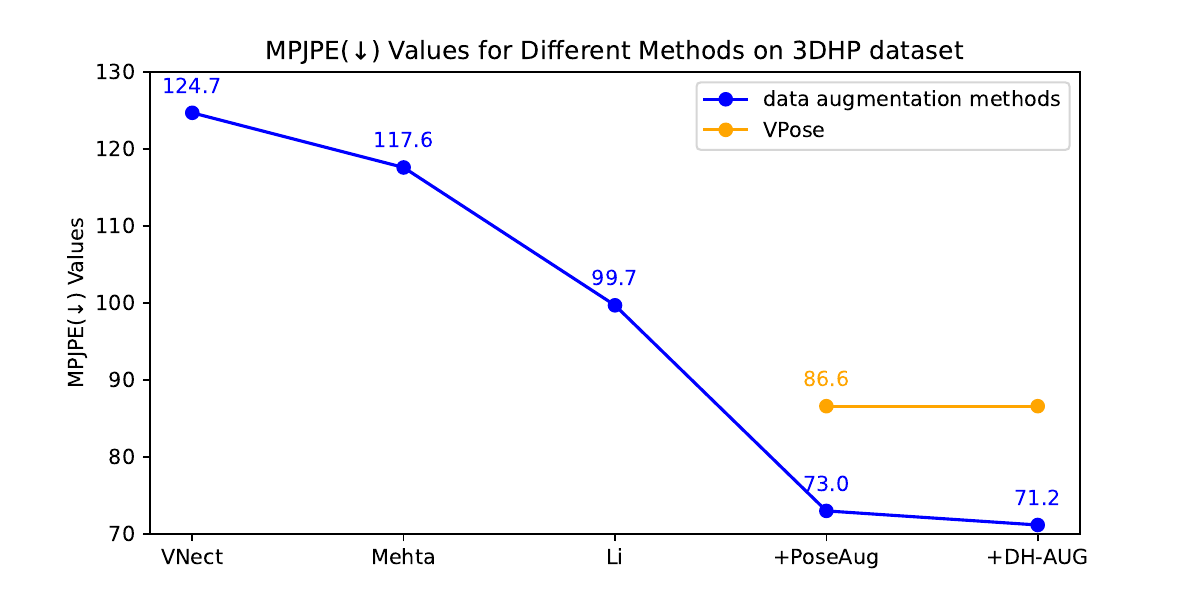}
  \caption{
\new{MPJPE of different methods on 3DHP Dataset. The baseline has achieved good results on this dataset, and data augmentation based on the baseline can also make significant progress.}  }
  \label{3dpose2}
\end{figure*}

\begin{figure*}[t]
  \centering
  \includegraphics[width=1.0\linewidth]{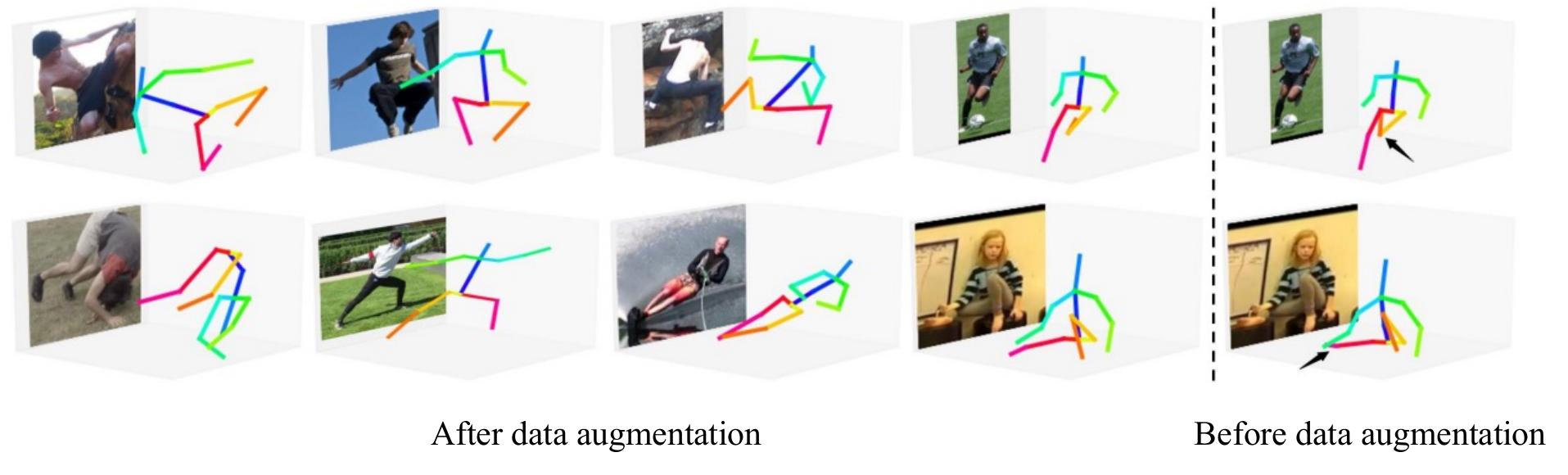}
  \caption{
\new{Example 3D pose estimations from LSP, MPII, 3DHP and 3DPW. Results of PoseAug~\cite{gong2021poseaug} are shown in the left four columns. The rightmost column shows the the results of Baseline—VPose~\cite{pavllo20193d}  trained without PoseAug. Errors are highlighted by black arrows.}  }
  \label{3dpose_vis}
\end{figure*}


\new{\textbf{Dataset.}
For the task of 3D human pose estimation, commonly employed datasets include Human3.6M~\cite{ionescu2013human3} and MPI-INF-3DHP Dataset~\cite{mehta2017monocular}. In the subsequent experimental analysis, we assess the impact of data augmentation methods in the domain of 3D human pose estimation on both the H3.6M and 3DHP datasets. Both datasets, H3.6M and 3DHP provide a substantial collection of images captured from diverse environments, accompanied by meticulous 3D keypoint annotations. This furnishes a solid foundation for evaluating the effectiveness of data augmentation methods in the realm of 3D human pose estimation.}

\textbf{3D human pose estimation experimental analysis.}
Based on Tables~\ref{tab:3dpose1} and~\ref{tab:3dpose2}, we can analyze the performance of different methods for 3D human pose estimation. \new{As shown in Figure~\ref{3dpose1} and~\ref{3dpose2}, we can see that with data augmentation methods, we can get better results. Figure~\ref{3dpose_vis} shows more accurate 3d human pose estimation with augmented data.}The evaluation metrics include MPJPE (Mean Per Joint Position Error), PCK (Percentage of Correct Keypoints), and AUC (Area Under the Curve). In Table~\ref{tab:3dpose1}, we can see that the data augmentation methods, Li, VPose + PoseAug~\cite{gong2021poseaug}, and VPose + DH-AUG~\cite{huang2022dh} show improvements in MPJPE compared to the non-data augmentation methods. In Table~\ref{tab:3dpose2}, which evaluates methods on the 3DHP dataset, among the data augmentation methods, Li, VPose + PoseAug~\cite{gong2021poseaug}, and VPose + DH-AUG~\cite{huang2022dh} demonstrate improvements in MPJPE compared to the non-data augmentation methods. Based on the provided information, we can conclude that VPose~\cite{pavllo20193d} is one of the better-performing methods for 3D human pose estimation in both datasets. VPose + DH-AUG also demonstrates good results, especially on the 3DHP dataset.

\subsection{Pedestrian Detection}

Pedestrian detection is the task of identifying and locating people within images or video frames, predominantly used in autonomous vehicle systems and urban surveillance. This task is crucial for ensuring pedestrian safety, as accurate and rapid detection of pedestrians enables vehicles or monitoring systems to react appropriately to avoid accidents. Modern pedestrian detection systems leverage deep learning models to distinguish pedestrians from various backgrounds and under different lighting conditions. The primary challenge here is to minimize false positives and negatives, ensuring reliable detection in diverse and often unpredictable urban settings.

In the pursuit of enhancing pedestrian detection models, a diverse array of data augmentation methods proves instrumental. Leveraging image-level perturbations, such as style transfer~\cite{michaelis2019benchmarking}, cutout~\cite{Cygert2020TowardRP,chi2020pedhunter}, and random erasing, introduces essential variability, allowing models to adapt to different visual conditions. Meanwhile, human-level perturbations focus on altering pedestrian shapes~\cite{chen2021shape}. It can involve techniques such as geometric transformations, shape warping, or body part swapping. By altering the pedestrian's shape, the model learns to recognize pedestrians across different body proportions, poses, and articulations.
Data recombination methods, including copy-paste techniques~\cite{dwibedi2017cut,li2021cutpaste, Deng2022ImprovingCO}, contribute to dataset diversity by rearranging image components. Furthermore, the integration of virtual data through graphics engine-based generation~\cite{Cheung2017MixedPedsPD,nie2022synposes, Lu2022PedestrianDF,nilsson2014pedestrian} introduces controlled variations, expanding the model's exposure to diverse scenes and environmental conditions. In the realm of generative approaches, utilizing generative models like GANs~\cite{Zhi2021PoseGuidedPI,zhang2020deep,liu2020novel} enables the generation of synthetic pedestrian images. This not only broadens the dataset but also empowers models to discern pedestrians across a spectrum of appearances, enhancing their generalization capabilities.

\begin{table*}
    \centering
    \begin{tabular}{|c|c|c|c|c|} \hline
            Methods&$MR^{-2}$ (↓)&AP@0.5 (↑)&AP@0.5:0.95 (↑)&JI (↑)\\ \hline
        \multicolumn{5}{|c|}{on Faster R-CNN~\cite{Ren2015FasterRT}}\\ \hline
        Baseline& 50.42 &84.95 &- &- \\ \hline
        Mosaic~\cite{bochkovskiy2020yolov4} &43.71 &85.21 &52.66 &78.35 \\
        RandAug~\cite{cubuk2018autoaugment} &42.17 &87.48 &53.19 &80.40 \\
        SAutoAug~\cite{chen2021scale} &42.13 &87.64 &53.35 &80.39 \\
        SimCP~\cite{ghiasi2021simple} &41.88 &87.36 &53.36 &79.53 \\
        CrowdAug~\cite{deng2023improving} & 40.21 &88.61 &54.88 &81.41 \\ \hline
        \multicolumn{5}{|c|}{on RetinaNet~\cite{lin2017focal}}\\ \hline
        Baseline &63.33 &80.83 &- &- \\ \hline
        Mosaic~\cite{bochkovskiy2020yolov4} &52.53 &82.95 &48.87 &75.60 \\
        RandAug~\cite{cubuk2018autoaugment} &50.25 &83.94 &49.77 &76.58 \\
        SAutoAug~\cite{chen2021scale} &50.21 &84.02 &49.85 &76.80 \\
        SimCP~\cite{ghiasi2021simple} &50.01 &84.12 &50.05 &77.02 \\
        CrowdAug~\cite{deng2023improving} & 47.35 &85.29 &51.84 &77.79\\ \hline
    \end{tabular}
    \caption{Pedestrian detection performance comparison of methods with Data Augmentation on CrowdHuman val set. Results are in percentage (\%). We compared data augmentation methods based on the Faster R-CNN and RetinaNet backbones.}
    \label{tab:pd}
    \vspace{-5mm}
\end{table*}

\begin{figure*}[t]
  \centering
  \includegraphics[width=0.8\linewidth]{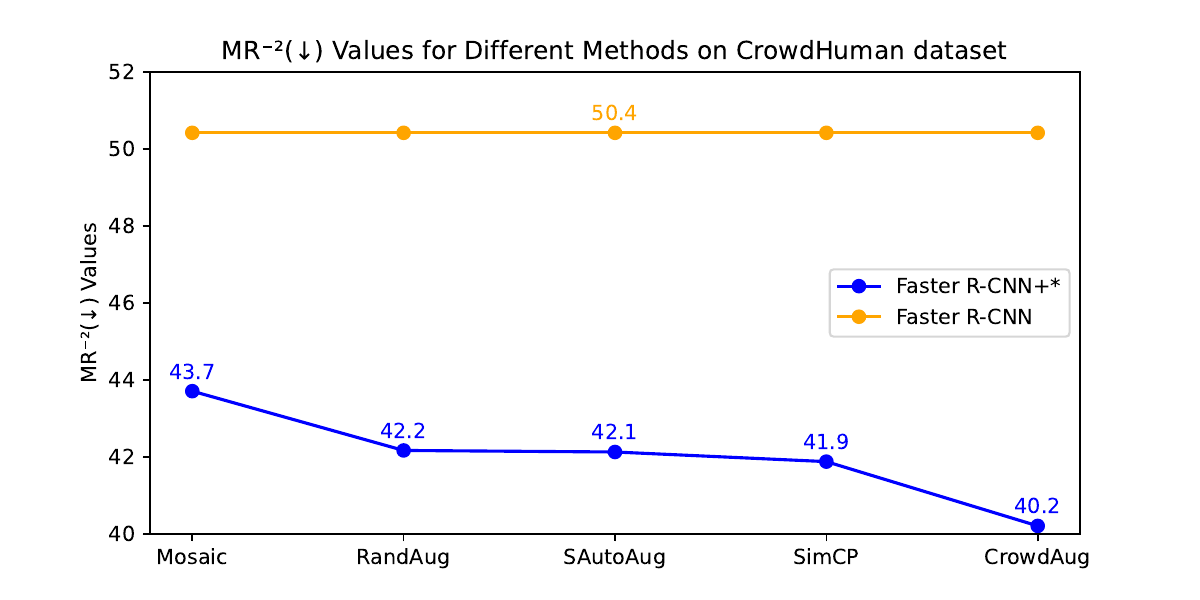}
  \vspace{-3mm}
  \caption{
\new{$MR^{-2}$ of different data augmentation methods with baseline Faster R-CNN on CrowdHuman Dataset. All data augmentation methods show significant performance improvements over baseline.}  }
  \label{pd1}
\end{figure*}

In essence, these augmentation strategies collectively contribute to a holistic training approach, enriching the dataset with realistic variations and bolstering the model's adaptability to the intricacies of real-world pedestrian scenarios.

\begin{figure*}[t]
  \centering
  \includegraphics[width=0.8\linewidth]{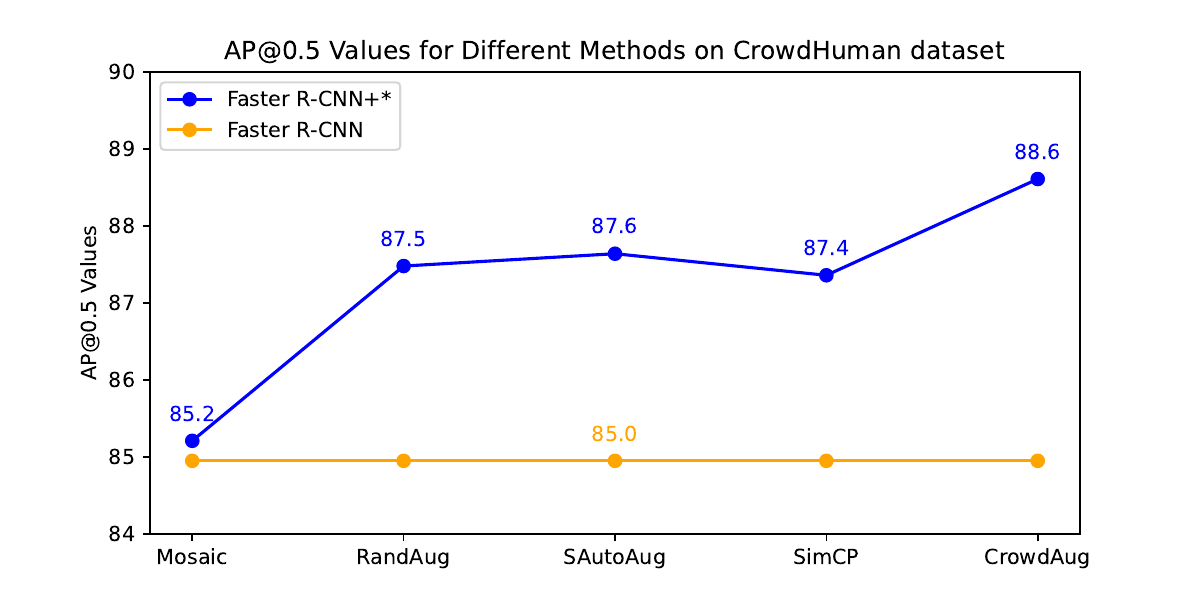}
  \vspace{-4mm}
  \caption{
\new{AP@0.5 of different data augmentation methods with baseline Faster R-CNN on CrowdHuman Dataset.}   }
  \label{pd2}
\end{figure*}


\begin{figure*}[t]
  \centering
  \includegraphics[width=0.7\linewidth]{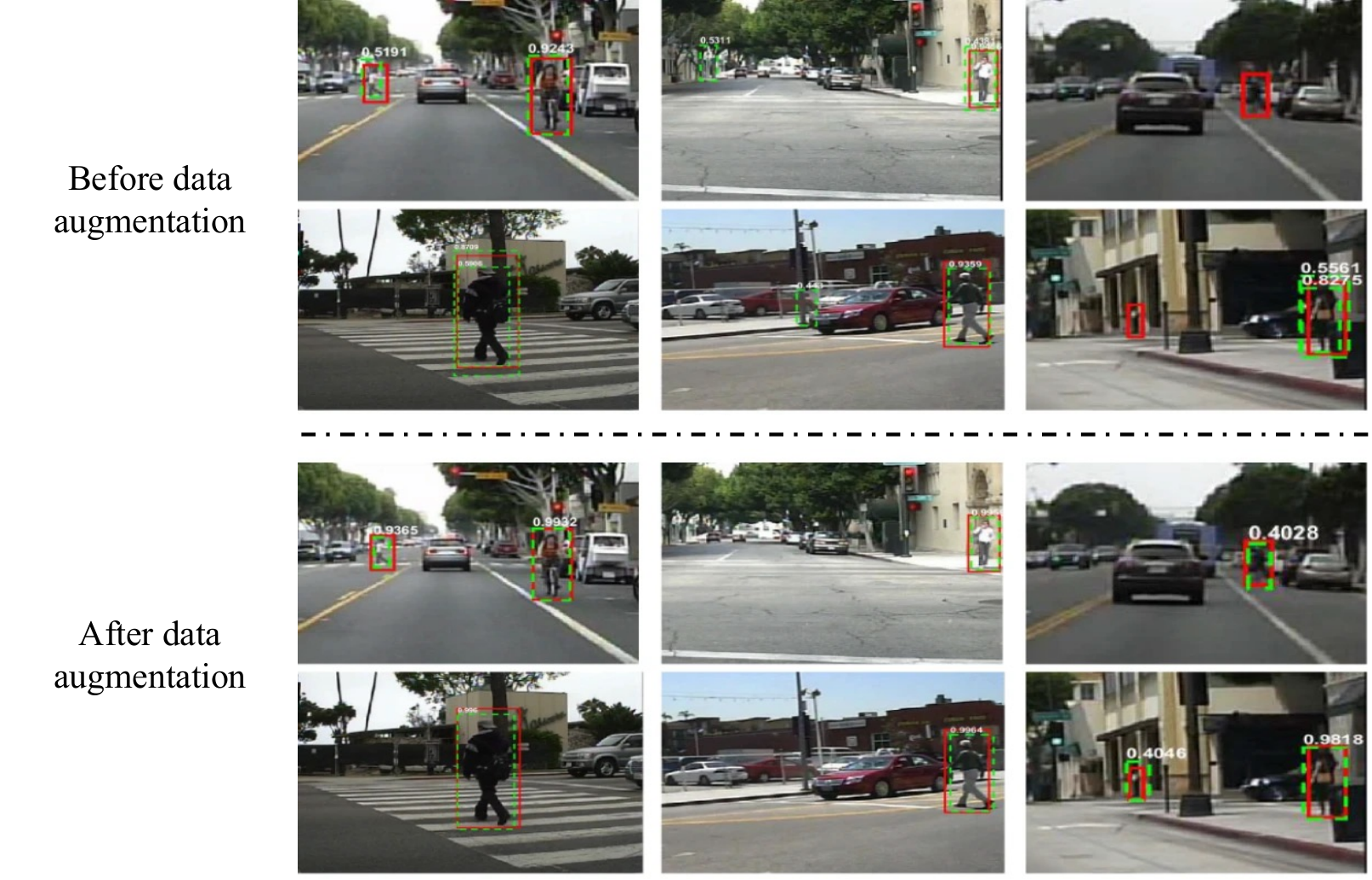}
  \caption{
\new{Comparison of qualitative results about whether applying STDA~\cite{chen2021shape} for data augmentation. Red boxes are ground-truths. Green dotted boxes are detection results.}   }
  \label{detection_vis}
\end{figure*}


\new{\textbf{Dataset.}
Commonly used datasets for Pedestrian Detection tasks include CrowdHuman~\cite{shao2018crowdhuman} and the INRIA Person Dataset~\cite{dalal2005histograms}. To assess the effectiveness of various data augmentation methods in the context of Pedestrian Detection, we opt to conduct our evaluation on the CrowdHuman dataset. Comprising approximately 15,000 images, each densely populated with pedestrian instances, the CrowdHuman dataset presents a diverse range of challenges, including different perspectives, various occlusions, and complex scenarios involving intersecting pedestrians. The dataset offers rich annotation information, facilitating the evaluation of algorithmic performance. This choice of dataset allows for a comprehensive validation of data augmentation methods in the domain of Pedestrian Detection.}

\textbf{Pedestrian detection experimental analysis.}
Table~\ref{tab:pd} compares the performance of different data augmentation methods on pedestrian detection accuracy using Faster R-CNN~\cite{Ren2015FasterRT} and RetinaNet~\cite{lin2017focal} models on the CrowdHuman validation dataset. \new{Figure~\ref{pd1} and~\ref{pd2} clearly demonstrates the superiority of data augmentation methods over the baseline in detecting pedestrians, as evidenced by their ability to detect pedestrians that were missed by the baseline, as shown in Figure~\ref{detection_vis}.} The evaluation metrics used in Table~\ref{tab:pd} include $MR^{-2}$ (Mean Recall at 2 False Positives per Image), AP@0.5 (Average Precision at 0.5 Intersection over Union threshold), AP@0.5:0.95 (Average Precision at 0.5 to 0.95 Intersection over Union threshold range), and JI (Jaccard Index). For both Faster R-CNN~\cite{Ren2015FasterRT} and RetinaNet~\cite{lin2017focal} models, we observe that the baseline model has the highest $MR^{-2}$ value, indicating a high percentage of recall at 2 false positives per image, but lower AP and JI values. On the other hand, all data augmentation methods improve the AP and JI values, indicating better precision and overlap between predicted and ground truth bounding boxes. Among the data augmentation methods, CrowdAug~\cite{deng2023improving} consistently outperforms other methods in terms of all evaluation metrics for both Faster R-CNN~\cite{Ren2015FasterRT} and RetinaNet~\cite{lin2017focal} models. SimCP~\cite{ghiasi2021simple} and SAutoAug~\cite{chen2021scale} follow closely behind CrowdAug~\cite{deng2023improving}, demonstrating competitive performance across evaluation metrics. RandAug~\cite{cubuk2018autoaugment} and Mosaic~\cite{bochkovskiy2020yolov4} also show improvements in AP and JI values compared to the baseline, although the improvement is relatively small. In conclusion, Table~\ref{tab:pd} highlights the effectiveness of data augmentation methods in improving the accuracy of pedestrian detection models on the CrowdHuman validation dataset, with CrowdAug~\cite{deng2023improving}, SimCP~\cite{ghiasi2021simple}, and SAutoAug~\cite{chen2021scale} being the most effective methods.

\begin{figure*}[t]
  \centering
  \includegraphics[width=0.9\linewidth]{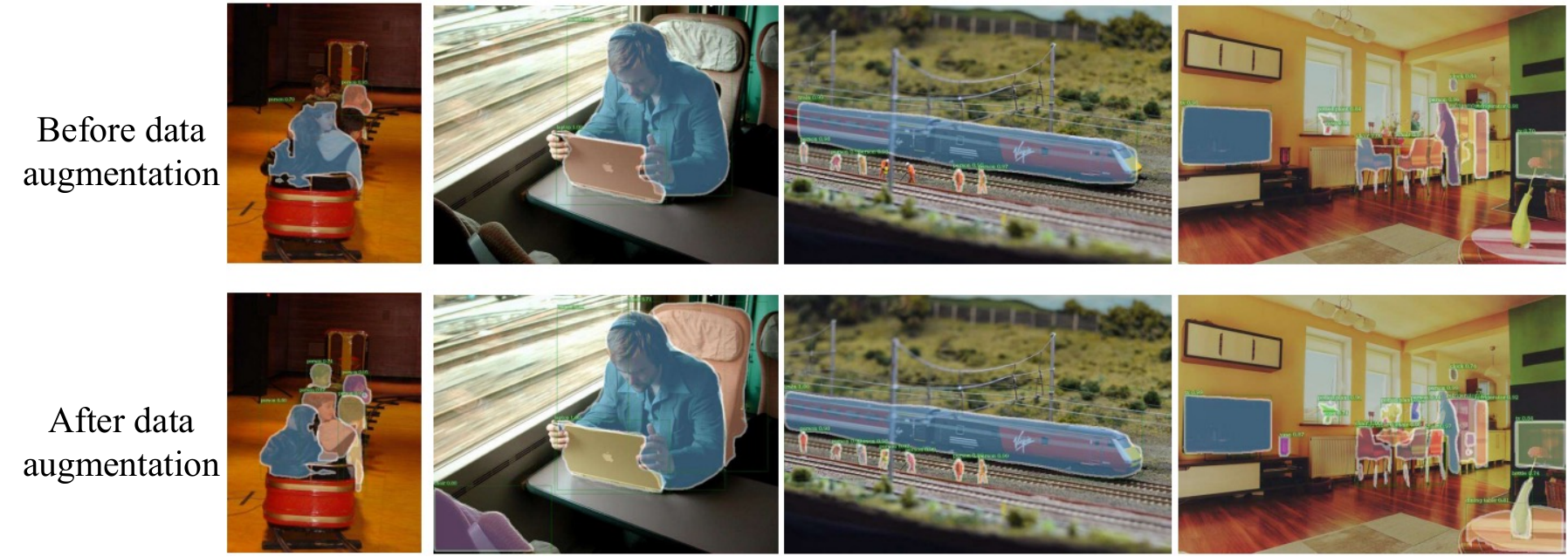}
  \caption{
\new{Instance segmentation result of vanilla Mask R-CNN~\cite{he2017mask} vs. Mask R-CNN trained with InstaBoost~\cite{fang2019instaboost} (bottom). InstaBoost guarantees finer instance segmentation results.}   }
  \label{parsing_vis}
\end{figure*}

\subsection{Human Parsing}

Human parsing refers to the process of segmenting a human image into multiple parts or regions, typically labeling each pixel with a category like head, arms, torso, or legs. This task is vital in applications such as augmented reality, fashion analysis, and advanced human-computer interactions. By understanding the spatial arrangement of different body parts, human parsing algorithms can provide detailed insights into human posture and attire, enabling personalized recommendations in fashion retail or accurate gesture recognition in interactive systems. The complexity of human parsing arises from the diversity in human poses, clothing styles, and body shapes, requiring sophisticated algorithms that can generalize well across varied scenarios.

In human parsing tasks, data augmentation methods similar to rotation and scale adjustments, which are common in image-level perturbation, can also be applied. However, specific data augmentation methods for human parsing are primarily image-level recombination, which includes image background replacement and copy-paste techniques. These methods are crucial for creating diverse training samples with varied backgrounds and compositions, leading to improved model performance and generalization. \new{Through data augmentation methods, we can segment the people in the image more accurately, as shown in Figure~\ref{parsing_vis}.}
The scarcity of dedicated data augmentation methods for human parsing tasks has resulted in limited experimentation. The absence of standardized datasets across related studies contributes to the challenge of comparing results. The use of different datasets by various research articles introduces inconsistencies and impedes direct result comparisons.

\new{\textbf{Dataset.}
The objective of Human Parsing tasks is to segment human body images into distinct semantic parts such as hair, clothing, and skin, thereby facilitating data augmentation. Datasets like the LIP (Look into Person) Dataset~\cite{liang2018look} and COCO~\cite{lin2014microsoft} are commonly used for training and evaluating Human Parsing algorithms. The LIP Dataset encompasses over 50,000 human body images, spanning diverse scenes and poses, and provides detailed annotations covering 50 different body parts. Similarly, the COCO dataset encompasses a multitude of object categories, including the human body, with a substantial number of annotated images suitable for human parsing tasks.}

\section{Future Work}
\label{sec4}

\begin{figure*}[t]
  \centering
  \includegraphics[width=1.0\linewidth]{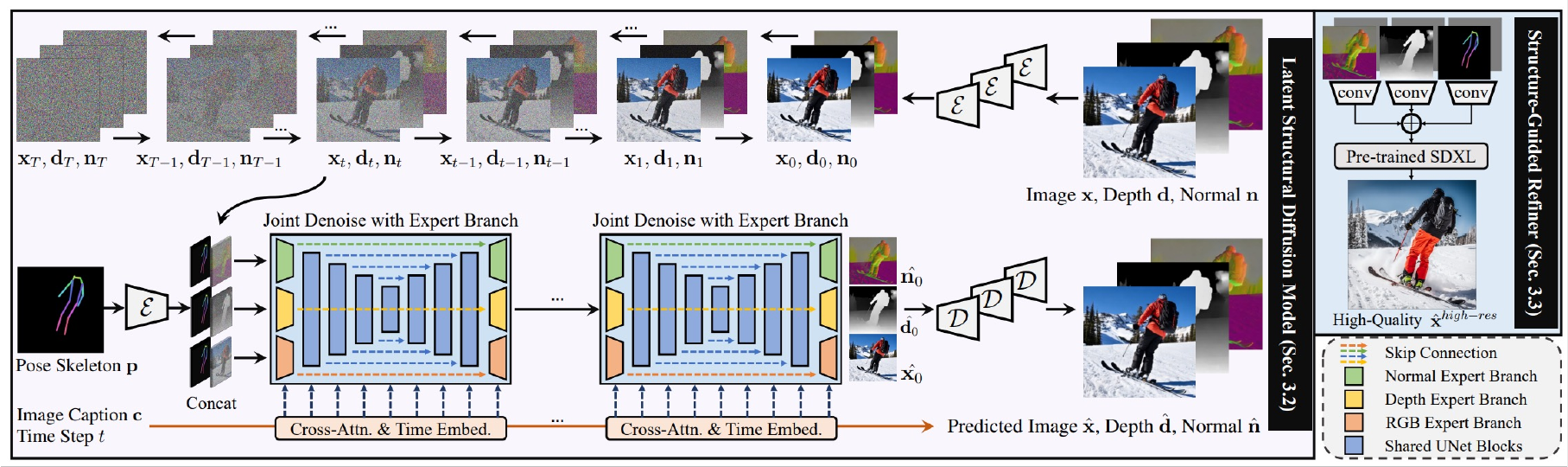}
  \caption{
    Visualization of using diffusion model~\cite{liu2023hyperhuman} for data generation using pose and depth as conditions.
  }
  \label{future_gen}
\end{figure*}

The future of leveraging data augmentation techniques in human-centric computer vision appears to be a promising avenue to address the challenges posed by overfitting or lack of training data in deep convolutional neural networks.
As these networks require extensive datasets to learn effectively, data augmentation can play a critical role in artificially expanding the dataset size and introducing a diverse range of variations in training samples.
This process helps in simulating real-world scenarios where changes in lighting, orientation, and background are common.
Looking forward, advancements in data augmentation are likely to focus on generating more realistic and complex augmentations that closely mimic real-world variations. This could involve integrating advanced generative models, like Generative Adversarial Networks (GANs)~\cite{creswell2018generative,goodfellow2020generative}, and Latent Diffusion Models (LDMs)~\cite{rombach2022high,pinaya2022brain} to create lifelike, varied training samples.
Moreover, the development of task-specific augmentation techniques, specially tailored for specific human-centric tasks such as human pose estimation or person ReID, will be crucial. These specialized augmentations would account for human-specific characteristics and movements, thereby enhancing the robustness and accuracy of these tasks.

\subsection{\textbf{Data Generation with Diffusion Models}}

Among the above future directions, the most promising one is to leverage the current powerful generative models, especially the pre-trained Latent Diffusion Models (e.g., Stable Diffusion~\cite{rombach2022high}) to generate human data for the human-centric vision for data augmentation.
Latent Diffusion Models (LDMs) like Stable Diffusion operate by gradually transforming a random noise distribution into a coherent image representation in a latent space. This process is guided by learned priors, allowing the generation of high-quality, diverse images. In the context of human-centric computer vision, these models can be adapted to generate human images with specific attributes, such as poses or appearances.

\textbf{Person Re-Identification (ReID):}
In person ReID, the model must recognize individuals across different scenes and camera angles. A controllable diffusion model can be used to generate images of the same person in various outfits, poses, and lighting conditions, as well as from different camera perspectives.
By inputting specific attributes or features (like clothing color, type, or individual physical characteristics) into the model, it can produce a diverse set of images that simulate different scenarios in which a person might be captured by surveillance systems. This helps in training ReID models to be more robust in identifying individuals despite changes in appearance or context.

\begin{figure*}[t]
  \centering
  \includegraphics[width=1.0\linewidth]{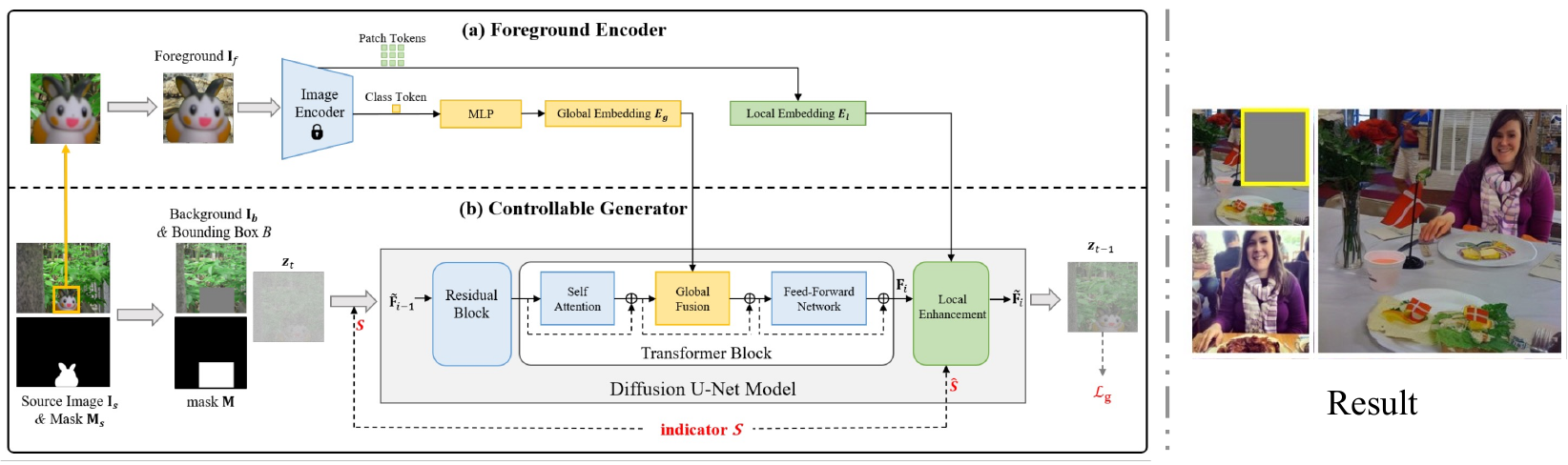}
  \caption{
    Visualization of using diffusion model~\cite{Zhang2023ControlComCI} for data recombination.
  }
  \label{future_recom}
\end{figure*}

\textbf{Pose-Guided Synthesis:}
Human pose estimation requires accurately identifying the position and orientation of various body parts. The diffusion model can be used to generate images of humans in a multitude of poses, from common to rare or challenging ones, in different environments, as shown in Figure~\ref{future_gen}.
Using models like ControlNet~\cite{zhang2023adding} or HyperHuman~\cite{liu2023hyperhuman}, which demonstrates proficiency in pose-guided image synthesis, we can generate human images in various poses.
This is particularly useful for tasks like human pose estimation, where diverse, accurately posed human figures can enrich the training data.
By inputting desired pose parameters, the model can synthesize human figures that match these poses, providing a rich and varied dataset for training human pose estimation models.
By training on these augmented datasets, human pose estimation models can achieve greater accuracy and flexibility in real-world applications, where human poses can vary greatly.

\textbf{Human Parsing:}
For tasks like human parsing, LDMs can be employed to create images based on parsing maps. These maps delineate different human body parts or clothing items, allowing the generation of images with a wide range of appearances and configurations.
This approach ensures diversity not only in poses but also in the representation of clothing and body types.
The model can be controlled to create complex scenarios where clothing and body parts are partially occluded or overlapping, a common challenge in real-world images. By training on these augmented datasets, human parsing models can learn to more effectively distinguish and segment various parts of the human body under diverse conditions.

\textbf{Pedestrian Detection:}
In pedestrian detection, the challenge lies in identifying individuals in a variety of urban settings and conditions. The diffusion model can create images of pedestrians in different urban landscapes, under various weather conditions, and during different times of the day.
The model can also simulate challenging scenarios, such as crowded scenes or pedestrians partially obscured by objects like vehicles or street furniture. Training on these augmented datasets enables pedestrian detection models to become more adept at recognizing pedestrians in complex and dynamic urban environments.

The generated data can be seamlessly integrated into existing training pipelines. By augmenting the dataset with a wide range of synthetic images, the model’s ability to generalize and perform accurately on real-world data is significantly enhanced. This approach also helps in addressing data scarcity issues, especially in domains where collecting extensive real human datasets is challenging or ethically problematic.

In summary, leveraging pre-trained Latent Diffusion Models for data augmentation in human-centric vision tasks holds immense potential.
By leveraging a controllable diffusion model, each of these tasks can benefit from a rich, diverse, and tailored dataset that addresses specific challenges inherent to that task. This approach not only enhances the robustness and accuracy of models in these areas but also significantly contributes to the advancement of human-centric computer vision technologies as a whole.

\subsection{Data Perturbation and Recombination with Diffusion Models}

Apart from generating the augmented images directly, data perturbation and recombination can leverage the current existing data to create more diverse data.
However, the previous methods that perform perturbation and recombination mainly use simple copy-paste and inpainting techniques, which produce artifacts that may harm the model training.
With the current powerful generative models like diffusion models, we can enable realistic and plausible data perturbation, and recombination, including background/foreground composition, human switching, and perturbation.

\textbf{Background/Foreground Composition}~\cite{zhang2023controlcom,zhang2023text2layer}:
The diffusion model, as shown in Figure~\ref{future_recom}, can be used to seamlessly integrate foreground human subjects into a variety of background scenes. This involves more than just superimposing figures onto backgrounds; it requires an understanding of lighting, perspective, and environmental context to make the composition realistic.
In tasks like pedestrian detection or person ReID, this technique helps create scenarios where subjects are placed in diverse environments, under different lighting conditions, and from various camera angles, greatly enhancing the diversity of training data.

\textbf{Human Switching}~\cite{schmitz2012decomposing,ging2020effects}:
This involves replacing one human subject in an image with another while maintaining the integrity and realism of the original scene. The diffusion model can recognize and adapt to the original image's context, such as lighting, pose, and interaction with the environment, ensuring a realistic switch.
For person ReID, this technique can be particularly useful, as it allows the creation of varied instances of the same individual in different settings or different individuals in the same setting, aiding the model in learning to focus on identifying features of persons rather than the context.

\textbf{Human Perturbation}~\cite{zhang2023enhanced}:
Here, the model introduces subtle changes to human subjects or their surroundings. This can include altering aspects like clothing textures, colors, or even background elements. The key is to do this in a way that maintains the overall realism of the image.
In human parsing and human pose estimation, such perturbations can create a more robust dataset by introducing minor variations that a model might encounter in real-world scenarios, improving its accuracy and generalization ability.

\section{Conclusion}

In this survey, we have presented an in-depth analysis of data augmentation techniques in the context of human-centric vision tasks. Our exploration distinctly categorizes these techniques into data generation and perturbation, offering a clear framework for understanding their application in person ReID, human parsing, human pose estimation, and pedestrian detection. This work stands out as the first comprehensive survey specifically addressing data augmentation for human-centric vision, providing a structured overview and critical insights into the nuances of these methods.
Future research directions point towards the integration of advanced generative models like Latent Diffusion Models for creating more realistic and diverse training data. This approach shows promise in enhancing model performance across various human-centric tasks by generating tailored, contextually appropriate augmented data. Such advancements are expected to significantly mitigate challenges such as overfitting and data scarcity, marking a substantial step forward in the field.
Overall, this survey not only synthesizes the current state of data augmentation in human-centric vision but also paves the way for novel methodologies and applications. The insights provided here will guide future efforts in developing more robust, precise, and efficient human-centric vision systems.

\section{Declarations}

\subsection{Availability of data and material}

This work is a comprehensive survey focused on reviewing existing data augmentation methods within human-centric vision research.
Given the nature of this study, it primarily involves the analysis and synthesis of findings from previously published papers.
As such, this survey does not involve the generation of new datasets or the creation of original materials that would necessitate release or archiving.

\subsection{Competing interests}
The authors declare the following affiliations: Wentao Jiang, Yige Zhang, Shaozhong Zheng, and Si Liu are affiliated with Beihang University. Shuicheng Yan is affiliated with Skywork AI. Beyond these affiliations, the authors declare no competing financial interests or personal relationships that could have appeared to influence the work reported in this paper.

\subsection{Funding}

This research is supported in part by National Key R\&D Program of China (2022ZD0115502), National Natural Science Foundation of China (NO. 62122010, U23B2010), Zhejiang Provincial Natural Science Foundation of China under Grant No. LDT23F02022F02, Key Research and Development Program of Zhejiang Province under Grant 2022C01082.

\subsection{Authors' contributions}
The survey paper was a collaborative effort, where Wentao Jiang, Yige Zhang, and Shaozhong Zheng, as students, were instrumental in conducting the survey, synthesizing findings, and drafting the manuscript. Professors Si Liu and Shuicheng Yan significantly contributed to guiding the research direction, framework, and providing critical revisions to ensure the manuscript's intellectual integrity. All authors have approved the final manuscript and agreed to be accountable for all aspects of the work.

\bibliography{sn-bibliography}

\end{document}